\documentclass{article}
\usepackage{iclr2027_conference,times}   
\iclrfinalcopy   
\usepackage{booktabs}
\usepackage{amsmath,amssymb}
\usepackage{graphicx}
\usepackage{hyperref}
\definecolor{linkblue}{HTML}{355C7D}
\hypersetup{
  colorlinks=true,
  linkcolor=linkblue,
  citecolor=linkblue,
  urlcolor=linkblue
}
\usepackage{url}
\usepackage{xcolor}
\usepackage{adjustbox}
\usepackage{tikz}
\usepackage{booktabs}
\usepackage{multirow}
\usepackage{makecell}
\usepackage{array}
\usepackage{adjustbox}
\usepackage[table]{xcolor}
\usepackage{siunitx}
\usetikzlibrary{arrows.meta,positioning,fit,backgrounds,calc}

\newcommand{\nummodels}{6}
\newcommand{\medsave}{50\%}
\newcommand{\numtraps}{5}
\newcommand{\maxpricex}{8.8$\times$}
\newcommand{\maxlatx}{68$\times$}

\newcommand{\equivcells}{4/17}

\iclrfinalcopy

\title{You Cannot Pick a Provider From the Price List: Market-Aware Routing for Open-Weight LLM Inference}

\author{
  Liang He,
  Jingbo Wen,
  Yixiong Chen,
  Yue Yang,
  Qizhen Lan,
  Kangning Cui,
  Xilu Wang
  \\[0.8em]
  {\small
  \begin{tabular}{@{}c@{\quad}c@{\quad}c@{\quad}c@{}}
    University of Sydney &
    Johns Hopkins University &
    Stanford University &
    City University of Hong Kong
  \end{tabular}
  }
}

\begin{document}
\maketitle

\lhead{}
\renewcommand{\headrulewidth}{0pt}

\begin{abstract}
Existing LLM routers choose among models using static per-model costs. We show that open-weight inference markets introduce a second, largely ignored decision axis: after choosing a model, a client must still choose which provider serves it. Measuring live endpoints across \nummodels{} open models, competing providers, multiple task types, and three measurement waves, we find that provider choice cannot be inferred from the price list. 
The same model can vary sharply in quality, latency, availability, and price
across providers; higher-priced providers are consistently faster, but price
does not reliably predict quality or availability; and provider feasibility is
task-selective, with one deployment nearly normal on knowledge tasks but
catastrophically degraded on multi-step reasoning.
We formulate same-model provider selection as a price-taker market-aware routing problem. A simple measured-map policy routes to the cheapest provider that is both quality-equivalent and healthy, yielding matched-quality savings while avoiding degraded endpoints. 
Because the map drifts, we introduce \textsc{facet}, an online provider router that certifies per-(provider$\times$task) feasibility facets and fails safe to an anchor before serving uncertified arms. Across relaxed deployment assumptions, \textsc{facet} tolerates imperfect task
assignment and sparse feedback, while systematic evaluator bias exposes a
quality-signal trust boundary that can be mitigated with ground-truth probes
or audits.
Live provider runs further confirm that certification can move real traffic from a premium anchor to a substantially cheaper certified endpoint.
Our results suggest that market-aware LLM routing must measure not only which model to use, but also who serves it.
\end{abstract}

\section{Introduction}

LLM inference is becoming a market-mediated service. Open-weight models such as Llama, Qwen, DeepSeek, Mistral, and Gemma are no longer served by a single source: the same weights are hosted by many competing providers, through different quantization choices \citep{frantar2023gptq,lin2024awq}, kernels, batching and scheduling policies \citep{kwon2023vllm,agrawal2024sarathi}, timeout behavior, and reliability profiles. At the same time, client-side LLM routing has become a practical way to reduce cost by selecting a model per query \citep{ong2024routellm,somerstep2025carrot,wang2025mixllm,sakota2024flyswat,yue2024cascades,hu2024routerbench}. Existing routers therefore ask a natural question: given a query, should we send it to a cheaper model or a stronger model?

This paper argues that this question is incomplete. In an open-weight inference market, selecting the model does not determine service. After a model router chooses, say, Llama-3.3-70B, the client must choose which provider serves it. This provider axis is not a minor implementation detail. Across providers, the same model can differ in price, latency, reliability, and answer quality. A routing system that treats "the model" as a stable object therefore hides an important operational decision.

The central question is therefore what provider choice should be based on.
Our empirical claim is simple: \textbf{a provider cannot be chosen from the price list.}
One might expect higher price to buy higher quality, lower latency, or better
availability. Our measurements reveal a more selective relationship.
Higher-priced providers are consistently faster at benchmark scale, but price
does not reliably predict answer quality or availability. The cheapest
provider is sometimes degraded, while the most expensive provider is often
neither the most accurate nor the safest. Price therefore exposes part of the
cost--speed frontier, but it does not reveal whether an endpoint is feasible
for a particular task. The advertised quantization label is not enough either
\citep{li2024quantized}: two endpoints with the same label can differ sharply in accuracy. Separating a safe cheap provider from a degraded endpoint therefore requires
direct, task-conditioned measurement.
A key fact revealed by these measurements is that the provider axis is also not global: a provider is not simply safe or unsafe. We find task-selective failures: a deployment can be nearly normal on a knowledge benchmark, noticeably degraded on code, and catastrophically broken on multi-step reasoning. Provider rankings can even invert across tasks. This means that per-model measurement is insufficient; the routing map must be indexed by model, task, and provider. Moreover, the map ages; silent behavioral change in a served endpoint is a documented
hazard for API consumers. Across three measurement waves over 43 days, prices move rarely, yet the cheapest quality-equivalent healthy provider changes because of availability recovery, rate limiting, provider exit, and silent quality slips. 
Market-aware routing in today's open-weight market is therefore
measurement-driven routing under drifting provider feasibility, with price
dynamics that depend on sampling cadence and provider-pool depth.

\begin{figure}[t]
\centering
\includegraphics[width=0.94\textwidth]{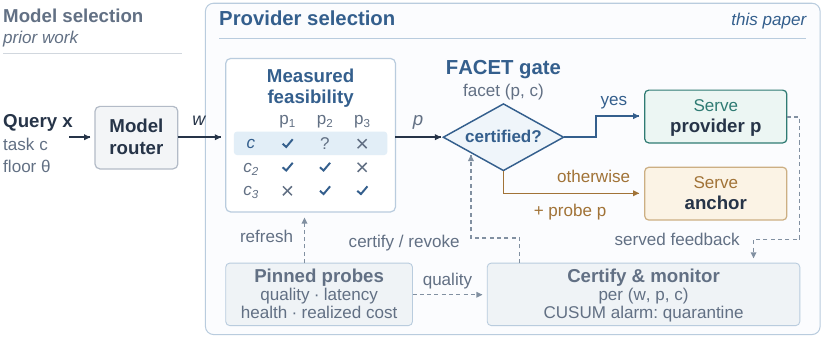}
\caption{The provider-selection layer. After an upstream router selects model
$w$, provider-pinned probes maintain a task-conditioned feasibility map.
\textsc{facet} serves a provider only when its $(p,c)$ facet is certified;
otherwise it falls back to the anchor while probing the candidate. Served
feedback supports continuous certification, revocation, and drift detection.}
\label{fig:system}
\end{figure}

We make this operational by turning live measurement into the routing loop shown in Figure~\ref{fig:system}: provider-pinned probes build a task-conditioned feasibility map, the router chooses the cheapest measured-equivalent healthy provider, and online feedback updates certificates as the map drifts. We later test how this design behaves when task assignment and quality feedback are imperfect rather than oracle-clean.
Following this measurement-to-routing loop, the paper makes four contributions.
\begin{enumerate}
\item \textbf{Provider selection as a missing routing axis.}
We formalize same-model provider selection as the routing problem that remains after an open-weight model has been chosen, under quality, health, latency, and cost constraints.
\item \textbf{Live measurement of provider heterogeneity beyond price.}
Using provider-pinned measurements across models, tasks, providers, and multiple
waves, we show that price contains a consistent speed signal but does not reveal
task-conditioned quality feasibility, which is both task-selective and
time-varying.
\item \textbf{Measured-map routing under drifting provider feasibility.}
We introduce a cheapest-equivalent-healthy router that converts live measurements into matched-quality savings, and evaluate how its decisions degrade as the provider map ages.
\item \textbf{Online feasibility certification beyond the clean setting.}
We formulate online provider selection as an inverse bandit and introduce
\textsc{facet}, which certifies per-(provider$\times$task) feasibility before
serving cheap endpoints. We separate the roles of certification and continuous
monitoring, test robustness to imperfect task assignment and feedback, and
identify ground-truth probes and audits as safeguards against systematic
evaluator bias; we also validate live certification and migration.
\end{enumerate}
\section{Setting and Measurement}
\label{sec:form}

We study the client-side decision that remains after a model has already
been chosen. For a query $x_t$, let $c_t$ denote its underlying task context.
The router need not observe $c_t$ directly; instead, it receives an observed
or inferred context signal $\hat{c}_t$, such as an application-provided label,
a lightweight classifier output, or request metadata. If a query cannot be
assigned confidently to a known context, it may be treated as an uncertified
"unknown" facet and fail safe. We initially evaluate the clean setting in
which task context is correctly identified, and relax this assumption in
Section~\ref{sec:robust}.

Suppose an upstream router or application selects an open-weight model
$w$. The remaining decision is to choose a provider
$p\in\mathcal{P}(w)$ serving it. Each provider exposes a
client-visible state
\[
s_p(t)=\bigl(p_p(t),\,\ell_p(t),\,h_p(t)\bigr),
\]
where $p_p(t)$ is the posted token price, $\ell_p(t)$ is measured latency,
and $h_p(t)$ is a health or availability signal. 
What is not directly known is the task-conditioned quality currently delivered
by the provider. We denote this latent quality by
\[
q_p(w,c,t),
\]
where $q_p(w,c,t)$ represents the expected task performance of provider $p$
for model $w$ under context $c$ at time $t$. A provider is feasible whenever
its quality clears the required task-conditioned floor.
Given a task-conditioned quality floor $\theta_{w,c_t}$ and, when applicable,
a latency budget $\ell_{\max}$, the ideal provider-selection problem is
\[
\min_{p\in\mathcal{P}(w)}
\mathrm{cost}_p(x_t,t)
\qquad
\text{s.t.}
\qquad
q_p(w,c_t,t)\ge \theta_{w,c_t},
\quad
h_p(t)\ge h_{\min},
\quad
\ell_p(t)\le \ell_{\max}.
\]
In the measured-map experiments, the quality floor is defined relative to the
best measured provider in each model--task cell,
\[
\theta_{w,c}=q_{\max}(w,c)-\delta,
\]
with $\delta=0.05$ in the primary setting.
When no latency SLA is specified, we take $\ell_{\max}=\infty$, recovering the
latency-unconstrained setting used in the primary experiments.
Appendix~\ref{app:latency}
evaluates explicit latency constraints. The measured-map router estimates the
task-conditioned quality constraint from provider measurements, while
\textsc{facet} later maintains that constraint online.

We obtain the measurements through a public multi-provider aggregator, pinning every request to a single provider and disabling
fallbacks. For each model--task--provider cell, we record accuracy, realized
cost, mean latency, and availability. Our light probes cover multi-step math,
extraction, sentiment classification, and code-output prediction, and we also
run GSM8K \citep{cobbe2021gsm8k}, MMLU \citep{hendrycks2021mmlu}, and HumanEval
\citep{chen2021humaneval} at benchmark scale. Appendix~\ref{app:measurement}
provides the full probing protocol and discusses probe-size limits, rate
limiting, token-budget artifacts, and aggregator/geographic effects.

This provider-selection layer is orthogonal to standard model routing:
RouteLLM, FrugalGPT, CARROT, and MixLLM
\citep{ong2024routellm,chen2023frugalgpt,somerstep2025carrot,wang2025mixllm,sakota2024flyswat,yue2024cascades}
choose which model answers a query, whereas our layer starts after that
decision and selects among providers serving the chosen model. We evaluate
their composition in Section~\ref{sec:router} and
Appendix~\ref{app:modelrouter}; broader comparisons with related routing,
pricing, and serving work appear in Appendix~\ref{app:positioning}.
The setting also differs from a classical reward bandit. Price, latency, and
client-visible health are observable before routing, while task-conditioned
quality is latent and must be estimated from probes or served-query feedback.
The resulting feasibility constraint,
$q_p(w,c_t,t)\ge\theta_{w,c_t}$, is therefore the unknown part of the routing
decision. We use the observed market state to identify attractive candidates
and spend measurement effort on determining whether those candidates may
safely serve traffic. Appendix~\ref{app:inverse-bandit} discusses this
observed-objective, latent-constraint view in relation to safe, constrained,
and change-detection bandits.
\section{The Provider Axis Is Real}
\label{sec:findings}

\begin{figure}[t]\centering
\includegraphics[width=0.8\textwidth]{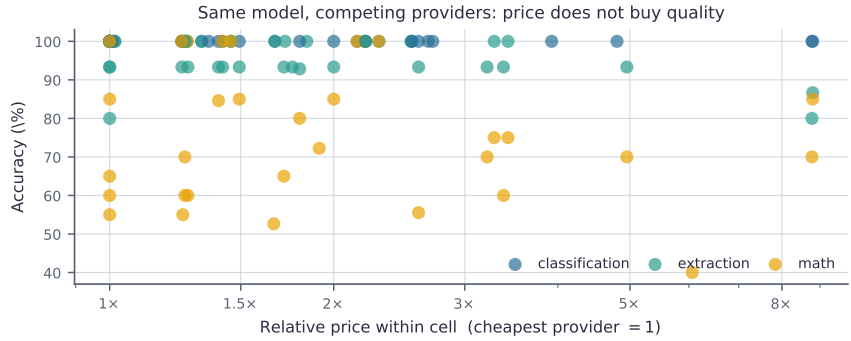}
\caption{Provider price does not predict same-model accuracy. Each point
represents a provider serving a fixed model on a fixed task, with price
normalized by the cheapest provider in the same cell. Providers with very
different prices often achieve comparable accuracy.}
\label{fig:scatter}
\end{figure}

We first ask whether provider choice remains meaningful once the model has
already been selected. If a model name were a stable service contract, then
providers serving the same weights should be largely interchangeable, and a
client could choose among them using price, latency, or a simple health signal.
The measurements show otherwise. Across the model--task cells in
Table~\ref{tab:main}, the same model can be offered at prices differing by up
to \maxpricex{} and latencies differing by up to \maxlatx{}, while accuracy can
differ by tens of percentage points on harder tasks. In \numtraps{} cells, the
cheapest provider falls below the quality floor. Provider choice therefore
remains consequential even after the model itself has been fixed.
More importantly, this heterogeneity has structure that matters for routing.
On easy cells, several providers reach the same or nearly the same accuracy
despite substantial price differences. On harder reasoning cells, the provider
spread widens and some endpoints fall well below the same-model provider pool.
Some cells therefore contain interchangeable providers and admit large savings,
while others contain degraded endpoints that must be excluded. Selecting the
cheapest measured-equivalent healthy provider yields a median saving of
\medsave{} relative to the premium provider at matched quality. Restricting the
analysis to the nine non-saturated cells with measurable cross-provider
quality differences still yields a $50.0\%$ median saving, compared with
$56.5\%$ over all cells (Appendix~\ref{app:nonsaturated}).
\begin{table}[t]
\centering
\small
\caption{\textbf{Provider dispersion across (model$\times$task) cells.}
Despite using the same model, providers differ substantially in accuracy,
price, and latency. The cheapest provider frequently matches higher-priced
alternatives, but may fall below the quality floor in some cells.}
\label{tab:main}

\renewcommand{\arraystretch}{1.02}
\setlength{\tabcolsep}{3.2pt}

\begin{adjustbox}{max width=\linewidth}
\begin{tabular}{@{}llrccccr@{}}
\toprule
&&&
\multicolumn{3}{c}{\textbf{Provider heterogeneity}}
&
\multicolumn{2}{c}{\textbf{Routing outcome}} \\
\cmidrule(lr){4-6}
\cmidrule(lr){7-8}

Model
& Task
& $|P|$
& Acc.\ range
& Price$\times$
& Lat.$\times$
& Route price
& Save \\
\midrule

& classification & 7 & 100\%--100\% & 2.3 & \textbf{68} & \$0.500 & 57\% \\
\cellcolor{gray!12}\emph{deepseek-chat-v3.1}
& extraction & 6 & 93\%--100\% & 1.8 & 34 & \$0.625 & 46\% \\
& math & 7 & 100\%--100\% & 2.3 & 24 & \$0.625 & 46\% \\
\midrule

& classification & 5 & 100\%--100\% & 2.5 & 4 & \$0.120 & 61\% \\
\cellcolor{gray!12}\emph{gemma-3-27b-it}
& extraction & 5 & 100\%--100\% & 2.5 & 8 & \$0.120 & 61\% \\
& math & 4 & 53\%--72\%\,$\dagger$ & 1.9 & 3 & \$0.305 & 0\% \\
\midrule

& classification & 5 & 100\%--100\% & 8.8 & 8 & \$0.025 & \textbf{89\%} \\
\cellcolor{gray!12}\emph{llama-3.1-8b-instruct}
& extraction & 5 & 80\%--93\%\,$\dagger$ & 8.8 & 10 & \$0.035 & \textbf{84\%} \\
& math & 5 & \textbf{56\%--85\%}\,$\dagger$ & 8.8 & 13 & \$0.220 & 0\% \\
\midrule

& classification & 9 & 100\%--100\% & 4.8 & 25 & \$0.265 & 79\% \\
\cellcolor{gray!12}\emph{llama-3.3-70b-instruct}
& extraction & 10 & 93\%--100\% & 5.0 & 6 & \$0.210 & \textbf{80\%} \\
& math & 11 & \textbf{40\%--75\%}\,$\dagger$ & 6.1 & \textbf{56} & \$0.265 & 79\% \\
\midrule

& classification & 4 & 100\%--100\% & 2.0 & 3 & \$0.375 & 50\% \\
\cellcolor{gray!12}\emph{llama-4-maverick}
& extraction & 4 & 93\%--93\% & 2.0 & 4 & \$0.375 & 50\% \\
& math & 4 & 80\%--85\% & 2.0 & 2 & \$0.375 & 50\% \\
\midrule

& classification & 4 & 100\%--100\% & 1.5 & 2 & \$0.172 & 14\% \\
\cellcolor{gray!12}\emph{mistral-small-3.2-24b-instruct}
& extraction & 4 & 93\%--100\%\,$\dagger$ & 1.5 & 3 & \$0.200 & 0\% \\
& math & 4 & 100\%--100\% & 1.5 & 4 & \$0.138 & 31\% \\

\bottomrule
\end{tabular}
\end{adjustbox}

\end{table}
Price alone does not distinguish safe from unsafe endpoints.
Figure~\ref{fig:scatter} shows both cheap providers that match the best endpoint
and cheap providers that lose substantial accuracy. Benchmark-scale latency
measurements, however, reveal a different pattern for speed: higher-priced
providers are consistently faster, even though price remains uninformative
about task-conditioned quality feasibility.
\begin{table}[t]
\centering
\small
\caption{\textbf{Benchmark-scale within-(model$\times$task) Spearman correlations}
between provider price and metrics. Price is strongly associated with
lower latency, but not with accuracy or availability.}
\label{tab:corr}
\begin{tabular}{lcc}
\toprule
Price vs. & Median Spearman & Interpretation \\
\midrule
Accuracy     & $+0.05$ & no stable price--quality relation \\
Latency      & $\mathbf{-0.61}$ & \textbf{higher-priced providers are faster} \\
Availability & $0.00$  & no stable price--availability relation \\
\bottomrule
\end{tabular}
\end{table}
Table~\ref{tab:corr} reveals a sharp distinction between speed and feasibility.
The median price--latency correlation is $-0.61$, with all $21/21$
benchmark-scale cells negative. By contrast, the corresponding correlations
with accuracy and availability are $+0.05$ and $0.00$, and neither exhibits a
stable relationship with price across alternative measurement-run choices.
Price can therefore inform the cost--speed trade-off, but cannot determine
whether an endpoint clears the task-conditioned quality requirement.
Appendix~\ref{app:latency} reports the complete per-cell analysis and
robustness checks.

A provider map is useful because the market contains both equivalence and
failure: equivalent providers expose price savings, while degraded endpoints
must be excluded. Concurrent cross-provider measurements
\citep{samemodelservice2026}, aggregate model leaderboards
\citep{chiang2024arena}, and platform-side derankers document this variation
but do not close the loop into task-conditioned routing under drift. We next
show that feasibility varies across tasks and over time.

\section{Feasibility Is Task-Selective and Drifting}
\label{sec:bench}

The previous section establishes that safe provider choice cannot be inferred
from price alone. We next ask whether a provider can be certified once at the
model level, or whether feasibility must be measured at a finer granularity.
We evaluate GSM8K, MMLU, and HumanEval across all measured provider endpoints,
yielding 93 endpoint-cells with $n{=}150$ for GSM8K and MMLU and
$n{=}100$ for HumanEval. Many same-model provider sets form tight quality
equivalence classes, which creates room for cheapest-equivalent routing.
However, the exceptions are task-selective. The same endpoint can remain
competitive on one workload while falling substantially below the provider
field on another. In the clearest case, a Llama-3.3-70B deployment is close to
the provider field on knowledge-oriented evaluation but becomes severely
degraded on multi-step reasoning. A single provider-level safe/unsafe label is
therefore too coarse; feasibility must be indexed by model, task, and provider.
The complete benchmark-scale results and the 5-point TOST equivalence analysis
are reported in Appendix~\ref{app:feasibility-drift} and
Appendix~\ref{app:tost}.

The map is also time-varying. We repeat the full measurement process 13 and
43 days after the initial collection. Posted prices remain mostly stable, yet
the selected route changes in $4/17$ comparable cells over days 0--13 and
$2/18$ over days 13--43. These changes are driven by availability recovery,
rate limiting, provider exit, and occasional quality degradation rather than
by price movement alone. The provider-level route transitions are reported in
Appendix~\ref{app:feasibility-drift}.
Finer-grained price monitoring further shows that the apparent stickiness of
list prices does not imply a fixed routing state. Over a 24-hour window at
15-minute resolution, deeper provider pools exhibited multiple re-elections
within a single day, while the shallower pools in our watch set remained
static. Moreover, prices observed through the aggregator reflect a flat contracted rate rather than the full time-varying market price, as confirmed by direct API testing against a first-party provider's published peak/off-peak schedule. Appendix~\ref{app:pricewatch} gives the complete analysis.

Provider feasibility therefore has two properties that matter for routing:
it must be task-conditioned and cannot be assumed to remain fixed.
The measured-map router addresses the first issue; once the map ages,
\textsc{facet} addresses the second through online feasibility certification.

\section{Measured-Map Routing}
\label{sec:router}

We next convert the measurement map into a routing policy. For each
model--task cell, the router selects the cheapest provider whose measured
accuracy is within 5 points of the best provider and whose availability
exceeds 90\%.
We treat these thresholds as deployment choices rather than tuned constants.
Across out-of-sample threshold sweeps, the measured-map below-floor rate is
$0.0$--$22.2\%$ and never exceeds cheapest- or premium-provider routing
(Appendix~\ref{app:theta}). The policy is intentionally simple: it tests
whether direct measurement can turn provider heterogeneity into a routing decision.

In-sample, the measured-map router achieves 94\% mean accuracy at
1.57$\times$ the cheapest-provider cost while keeping the below-floor rate at
0\%. This is close to the 95\% accuracy of the single-best oracle, but at far
lower cost than the premium policy, which pays 3.67$\times$ the cheapest
provider cost and still sends 17\% of cells below the quality floor. Pure
cheapest-provider routing reduces cost to 1.0$\times$, but increases the
below-floor rate to 28\%. The full comparison, including the random baseline,
is reported in Appendix~\ref{app:offline-routing}.
These savings use the five-point margin on measured accuracies, raising the
question of how much of the cost advantage remains once sampling uncertainty
is taken into account. On a 14-cell benchmark subset, requiring a one-sided
95\% confidence interval to support the same five-point margin reduces median
saving from $55\%$ to $46\%$. Importantly, simply selecting the most accurate
provider already saves $45\%$ relative to the premium provider. Thus most of
the statistically supported saving comes from price--quality misalignment,
while the equivalence margin provides additional savings that require larger
samples to certify. Appendix~\ref{app:tost} reports the full confidence-based
analysis.
We then evaluate the same policy out of sample by selecting providers from an
earlier measurement wave and scoring them on a later wave. Under the primary
five-point quality margin and $90\%$ availability threshold, the selected
provider remains above the later-wave floor in all $9/9$ comparable cells.
This robustness is threshold-dependent: relaxing the availability threshold to
$80\%$ raises the OOS below-floor rate to $11.1\%$, while tightening the
quality margin to two points raises it to $22.2\%$. Appendix~\ref{app:theta}
reports the complete out-of-sample threshold sweep.

Provider-layer routing also retains independent value when model selection is
handled upstream. We stack it beneath a RouteLLM-style difficulty router that
first selects between an 8B and a 70B model. On GSM8K, the cheapest provider
is already safe, so the provider layer makes the same choice at the same cost.
On MMLU and HumanEval, however, cheapest-provider routing sends 55.9\% and
57.5\% of queries through below-floor endpoints. Provider-aware selection
reduces both rates to zero, while increasing cost only from \$328.3 to \$336.7
on MMLU and from \$319.7 to \$380.0 on HumanEval. Relative to always using the
dearest provider, it also retains most of the available cost saving. The
provider axis therefore does not disappear after model routing; it remains a
separate decision underneath it. Appendix~\ref{app:modelrouter} reports the
full stacked evaluation.
Cost minimization can nevertheless trade away latency. Across the $17$ cells
with at least two quality-feasible providers, the cheapest feasible endpoint
incurs a median $7.8\times$ p95-latency penalty relative to the fastest
feasible endpoint. Adding an explicit latency constraint exposes the
corresponding cost--speed trade-off: a 10-second SLA retains all $12/12$
evaluated cells at only $1.19\times$ the unconstrained routing cost, whereas a
1-second SLA leaves only $6/12$ cells feasible and raises cost to
$2.62\times$. Appendix~\ref{app:latency} reports the complete latency analysis
and deadline sweep.
The measured-map router therefore captures task-conditioned provider
heterogeneity, but it remains a snapshot-based policy. The next question is
how to preserve this safety once the underlying provider state begins to move.

\section{\textsc{facet}: Online Feasibility Certification}
\label{sec:facet}

A measured provider map can become stale as availability, provider membership,
or quality changes. Generic online bandits can adapt to these changes, but
their exploration rule is problematic here: a newly attractive arm may be a
catastrophically degraded provider, and exploring it means serving real user
traffic. Exact-replay experiments in Appendix~\ref{app:bandit} show that
bandits can fine-rank providers once feedback accumulates, but do not remove
this early-exposure risk.

\begin{figure}[t]\centering
\includegraphics[width=0.76\textwidth]{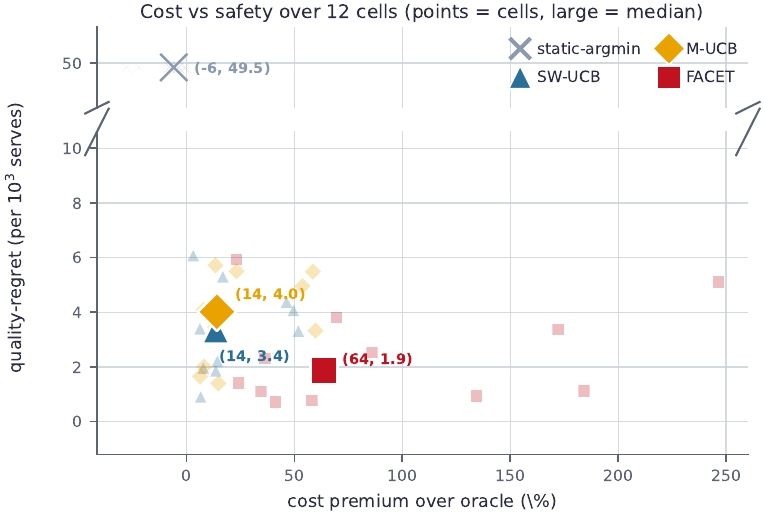}
\caption{Cost--safety trade-off for online provider selection.
\textsc{facet} pays a cost premium to avoid routing traffic to uncertified
low-cost providers, which cost-greedy policies may select prematurely.}
\label{fig:facet-pareto}
\end{figure}

\textsc{facet} exploits provider-market asymmetry: price, latency, and health
are observable, while task-conditioned quality is not. Each
(provider$\times$task) pair is a separate feasibility facet. A newly
attractive provider is therefore not served immediately; it first accumulates
quality evidence while traffic falls back to an anchor. In replay,
certification uses the 200 most recent labelled observations. Once at least
$n_{\min}=12$ observations are available, a facet is certified if its
windowed accuracy is within 8 points of the best same-task provider with at
least $n_{\min}$ observations and no more than 3 points below the task floor
$\theta$. After admission, the role shifts from certification to monitoring:
certified facets are continuously checked and quarantined when their quality
degrades. Replay uses these point-estimate rules; detector details and the
confidence-aware live rule are given in Appendices~\ref{app:guarantees},
\ref{app:detector}, and~\ref{app:live}.
Figure~\ref{fig:facet-pareto} frames \textsc{facet} as a safety--cost trade-off
rather than a pure cost-minimization policy, while
Table~\ref{tab:facet-ablation} isolates fail-safe anchoring and
task-conditioned certification by removing each under adversarial re-election.
In the anchoring ablation, S1 makes the cheapest arm a catastrophic mine from
the start; S2 lets a certified cheap arm slip mid-run; S3 reflects the
measured market without a cheap catastrophic mine. Removing the anchor
reproduces the optimistic policy's exposure, isolating the gain from the
never-serve-uncertified rule. In the task-certification ablation, a multi-task
stream interleaves GSM8K, MMLU, and HumanEval; a provider is safe on two tasks
but catastrophic on the third. A context-blind certifier pools evidence
across tasks, certifies globally, and then serves the unsafe task; per-facet
certification avoids this failure because evidence from one task does not
certify another. A simpler alternative to continuous monitoring is periodic
re-measurement: every $P$ rounds, probe all providers, refresh feasibility
estimates, and serve the cheapest provider whose latest measurement clears the
floor. This works when degradation is visible at a measurement round, but
creates a blind interval: if a certified provider slips immediately after
measurement, the router continues serving it until the next refresh.


\begin{table}[t]
\centering
\small
\caption{\textbf{Ablation experiments isolating fail-safe anchoring and
task-conditioned certification.}
Values report catastrophic exposure (below-floor served queries).}
\label{tab:facet-ablation}

\renewcommand{\arraystretch}{1.08}

\begin{tabular}{@{}lccc@{}}
\toprule
&
\multicolumn{3}{c}{\textbf{Catastrophic exposure}} \\
\cmidrule(lr){2-4}

Scenario
& \textsc{Facet} (full)
& No anchor
& Context-blind \\
\midrule

S1: Cheap mine from start
& $\mathbf{0.0\%}$
& 8.3\%
& 6.7\% \\

S2: Certified arm slips mid-run
& $\mathbf{0.4\%}$
& 5.6\%
& 3.2\% \\

S3: No cheap catastrophic mine
& 1.2\%
& 0.0\%
& --- \\

\bottomrule
\end{tabular}

\end{table}
\begin{table}[t]
\centering
\small
\caption{\textbf{Periodic re-measurement versus \textsc{facet}.}
A certified provider can slip between measurement rounds.
Periodic re-probing reduces exposure by probing more frequently, while
\textsc{facet} continuously monitors served facets and uses probes
only for certification (10 seeds).}
\label{tab:reprobe-main}

\renewcommand{\arraystretch}{1.08}
\setlength{\tabcolsep}{2.8pt}

\begin{tabular}{@{}cccc@{\hspace{8pt}}cccc@{}}
\toprule

\multicolumn{4}{c}{Periodic re-measurement}
&
\multicolumn{4}{c}{\textbf{\textsc{facet}}} \\
\cmidrule(lr){1-4}
\cmidrule(lr){5-8}

$P$
& Probe cost (\$)
& Below-floor
& Total cost (\$)
& $r$
& Probe cost (\$)
& Below-floor
& Total cost (\$) \\
\midrule

100
& 1740.2
& 3.70\%
& 2361
& 0.10
& 16.7
& $\mathbf{0.17\%}$
& 1255 \\

400
& 487.2
& $\mathbf{14.30\%}$
& $\mathbf{1089}$
& 0.25
& 22.9
& $\mathbf{0.67\%}$
& $\mathbf{1110}$ \\

1600
& 139.2
& 38.60\%
& 709
& 1.00
& 27.6
& $\mathbf{0.28\%}$
& 1130 \\

\bottomrule
\end{tabular}

\end{table}

Table~\ref{tab:reprobe-main} shows the gap between periodic
re-measurement and the \textsc{facet} system, but the comparison changes
both admission and monitoring at once. We therefore run a component control
that gives periodic re-measurement the same served-feedback monitor as
\textsc{facet}, leaving scheduled full-pool admission as the main difference.
The result matches the division of labor. When a previously safe
provider later degrades, the shared monitor is the relevant mechanism:
monitored periodic routing reaches $0.3$--$0.5\%$ below-floor exposure,
comparable to \textsc{facet} at similar total cost. Certification matters in
the complementary admission regime. When the cheapest provider is below the
floor from the outset, the never-serve-uncertified rule further reduces
cold-start exposure, while targeted certification lowers dedicated probe spend
by $8$--$23\times$ relative to scheduled full-pool probing. Certification
therefore protects admission and concentrates measurement on potentially
useful candidates, whereas continuous monitoring protects already-certified
providers under subsequent drift. Appendix~\ref{app:reprobe} reports the
40-seed component control and the full re-probing sweeps.

\begin{table}[t]
\centering
\small
\caption{\textbf{Robustness to imperfect task assignment.} Left: corruption of known
task labels. Right: OOD traffic treated as uncertified or mapped to the
nearest known facet. Values report catastrophic exposure (below-floor served
queries; 10 seeds).}
\label{tab:robust-task}
\setlength{\tabcolsep}{3pt}
\begin{tabular}{lcc|lcc}
\toprule
\multicolumn{3}{c|}{Label corruption (known tasks)}
& \multicolumn{3}{c}{OOD traffic} \\
\cmidrule(lr){1-3}
\cmidrule(lr){4-6}
Perturbation & \textbf{\textsc{Facet}} & Reference
& Perturbation & \textbf{\textsc{Facet}} & Reference \\
\midrule

$0\%$ error
& $\mathbf{0.00\%}$
& $6.72\%$ (context-blind)
& $10\%$ OOD
& $\mathbf{0.00\%}$
& $3.41\%$ (nearest facet) \\

$10\%$ error
& $\mathbf{2.23\%}$
& $6.72\%$
& $25\%$ OOD
& $\mathbf{0.00\%}$
& $1.03\%$ \\

$20\%$ error
& $\mathbf{3.54\%}$
& $6.72\%$
& $50\%$ OOD
& $\mathbf{0.00\%}$
& $0.24\%$ \\

$33\%$ error
& $\mathbf{4.72\%}$
& $\mathbf{6.72\%}$
& &
& \\

\bottomrule
\end{tabular}
\end{table}

The experiments above clarify why \textsc{facet} combines three separate
mechanisms. Fail-safe anchoring prevents newly elected but uncertified
providers from receiving user traffic, task-conditioned certification avoids
transferring safety evidence across workloads, and continuous monitoring
limits exposure when a previously certified provider later degrades.
Appendix~\ref{app:detector} compares the monitoring component with
non-stationary bandit baselines, while Appendices~\ref{app:drift} and
\ref{app:regret} examine its behavior under different drift patterns and the
additional cost of repeated re-certification.

This protection is not free. Whether the additional cost is worthwhile depends
on the application-level consequence of a below-floor answer. Against SW-UCB,
the break-even point is about $17\times$ the cost of an ordinary served query;
against a frozen map or periodic refresh, it is roughly one query-price or
less. Appendix~\ref{app:lambda} reports the full cost--safety analysis.
Because these signals may be imperfect in real deployments, we next test how
\textsc{facet} behaves when the assumptions underlying the preceding
experiments are relaxed.

\section{Beyond the Idealized Setting}
\label{sec:robust}

In deployment, task context may be wrong or unavailable, and quality feedback
may be sparse, noisy, or biased. We relax these assumptions progressively.
Missing or symmetric-noise evidence mainly triggers fallback, whereas
systematic bias exposes a harder certification boundary.

\textbf{Imperfect task assignment.} We first perturb task identity. In an interleaved GSM8K--MMLU--HumanEval stream, the task label observed by the router is randomly corrupted while safety is evaluated against the true task. Table~\ref{tab:robust-task} (Left) shows gradual degradation: below-floor exposure rises from $0\%$ to $4.72\%$ at $33\%$ label error, still below the $6.72\%$ exposure of context-blind certification that pools all tasks. Perfect task classification is therefore not required.
A separate challenge arises when queries fall outside the known task set. Reusing the nearest known facet can transfer safety evidence from an unrelated workload. \textsc{facet} instead treats unknown queries as uncertified and applies the fail-safe rule. Table~\ref{tab:robust-task} (Right) keeps catastrophic exposure at zero as OOD traffic reaches $50\%$, whereas forcing unknown queries into the nearest known facet produces unsafe serves. This assumes an OOD reject signal is available; open-set detection itself is a separate problem (Appendix~\ref{app:assumptions}).

\textbf{Imperfect quality feedback.}
The quality signal is fundamental. We first revisit symmetric judge
noise. Its near-zero below-floor exposure does not indicate evaluator
robustness: with $20\%$ random judge error, \textsc{facet} serves
$95$--$97\%$ of traffic from the anchor. Symmetric noise therefore
tests whether uncertainty induces conservative fallback.
We next consider an evaluator that systematically over-credits a weak
provider. With judged certification probes, exposure remains low under mild
bias but increases sharply as the bias strengthens. Scoring probes against
ground-truth answers protects cold-start certification. A provider that
degrades after admission remains dependent on served feedback, but a small
gold audit substantially reduces this exposure. These experiments identify
quality feedback as a trust boundary of \textsc{facet}, with mitigations
rather than an assumption of an arbitrarily reliable judge.
Appendix~\ref{app:assumptions} reports the bias sweep, sparse feedback,
and mitigation.

\textbf{Live deployment validation.} The experiments above are replay-based. We finally test whether certification and migration operate against live endpoints. We ran \textsc{facet} for $36$ hours on Llama-3.3-70B across 12 providers, interleaving fresh GSM8K, MMLU, and HumanEval queries. The run served 538 queries and issued 130 certification probes. Traffic began on a \$1.04/M-token anchor and moved to a certified provider at \$0.21/M tokens. Average served price fell from \$0.924/M to \$0.210/M after query 300, while served accuracy remained $88.3\%$.
Table~\ref{tab:live-facet} summarizes the run. Relative to the anchor, \textsc{facet} achieved a $63.7\%$ reduction in serving cost ($57.1\%$ after including probes). No provider quality collapse occurred during this window, so the experiment validates cold-start certification and live migration rather than slip detection. It also exposed deployment issues absent from replay---false quarantine at small sample sizes and wasted probes on unavailable endpoints---addressed by confidence-aware certification and availability backoff (Appendix~\ref{app:live}).

The live run and stress tests answer different questions. The live run shows
that certification can operate against real providers, while the stress tests
separate uncertainty from systematic error. Imperfect task assignment, sparse
feedback, and symmetric label noise mainly cause gradual degradation or
conservative fallback; systematic evaluator bias can instead corrupt the
quality signal itself. \textsc{facet} therefore still requires a maintained
anchor, some trustworthy quality evidence, and a way to assign or reject query
context. Ground-truth certification probes and occasional audits provide
practical safeguards when evaluator bias is a concern. In the limit of no
task signal and no trustworthy quality evidence, no certification-based method
can distinguish safe from unsafe providers.






\begin{table}[t]
\centering
\small
\caption{\textbf{Live \textsc{facet} deployment: 36-hour run on
Llama-3.3-70B across 12 providers.}
Validates cold-start certification and migration; no quality slip occurred
during the window.}
\label{tab:live-facet}

\renewcommand{\arraystretch}{1.04}
\setlength{\tabcolsep}{4pt}

\begin{tabular}{@{}lrlr@{}}
\toprule
Metric & Value & Metric & Value \\
\midrule

\rowcolor{gray!12}
\multicolumn{4}{@{}l}{\emph{Deployment scale}} \\
Served queries
& $538$
& Certification probes
& $130$ \\

Failed calls
& $4$
& Aggregate served accuracy
& $\mathbf{88.3\%}$ \\

\addlinespace[2pt]
\rowcolor{gray!12}
\multicolumn{4}{@{}l}{\emph{Serving behavior}} \\
Avg. price, queries 1--100
& $\mathbf{\$0.924 / M}$
& Avg. price, queries 301--500
& $\mathbf{\$0.210 / M}$ \\

Anchor price
& \$1.04 / M
& {}
& {} \\

\addlinespace[2pt]
\rowcolor{gray!12}
\multicolumn{4}{@{}l}{\emph{Cost outcome}} \\
Serving spend
& \$0.0635
& Probe spend
& \$0.0115 \\

Serving-cost reduction
& $\mathbf{63.7\%}$
& Total-cost reduction incl. probes
& $\mathbf{57.1\%}$ \\

\bottomrule
\end{tabular}

\end{table}
\section{Conclusion}

Open-weight inference markets make model choice incomplete. The same model served by competing providers can differ enough in quality, latency, availability, and price that a provider cannot be chosen from the price list. Live measurement shows that provider feasibility is task-selective and drifting, motivating both measured-map routing and \textsc{facet}'s online per-task certification. Our additional studies show that \textsc{facet}'s certification and monitoring
components address complementary failure modes, while systematic evaluator
bias defines an explicit quality-signal trust boundary that can be mitigated
with ground-truth probes or audits. Live deployment results demonstrate that certification can operate on real
provider traffic and move traffic from a premium anchor to a substantially
cheaper certified endpoint. These results suggest that market-aware LLM routing must decide not only which model to use, but also which provider currently has enough evidence to be trusted to serve it.
We discuss limitations and scope conditions in
Appendix~\ref{app:limitations}.

\bibliographystyle{iclr2027_conference}
\bibliography{references}
\appendix

\clearpage

\noindent{\large\scshape Appendix Contents}

\vspace{1.2em}

\newcommand{\appcontentsline}[3]{%
  \noindent
  \makebox[2em][l]{#1}%
  #2\ \dotfill\ \pageref{#3}\par
  \vspace{0.35em}
}

\appcontentsline{A}
{Limitations}
{app:limitations}

\appcontentsline{B}
{Related Work Positioning}
{app:positioning}

\appcontentsline{C}
{Detector-Core Comparison}
{app:detector}

\appcontentsline{D}
{Measurement Protocol and Caveats}
{app:measurement}

\appcontentsline{E}
{When Is the Safety Premium Worth Paying?}
{app:lambda}

\appcontentsline{F}
{Price--Latency Trade-offs and Latency-Constrained Routing}
{app:latency}

\appcontentsline{G}
{Run-to-Run Variability}
{app:variability}

\appcontentsline{H}
{Out-of-Sample Sensitivity to the Quality Floor and Health Threshold}
{app:theta}

\appcontentsline{I}
{Why Provider Selection Has an Inverse-Bandit Structure}
{app:inverse-bandit}

\appcontentsline{J}
{Drift-Shape Stress Tests}
{app:drift}

\appcontentsline{K}
{Measured-Map Routing Results}
{app:offline-routing}

\appcontentsline{L}
{Savings Beyond Saturated Cells}
{app:nonsaturated}

\appcontentsline{M}
{Periodic Re-Probing Baseline}
{app:reprobe}

\appcontentsline{N}
{Benchmark-Scale Feasibility and Multi-Wave Drift}
{app:feasibility-drift}

\appcontentsline{O}
{Assumption-Relaxation Experiments}
{app:assumptions}

\appcontentsline{P}
{Live Provider Deployment Details}
{app:live}

\appcontentsline{Q}
{Composition with an Upstream Model Router}
{app:modelrouter}

\appcontentsline{R}
{Exact-Replay Bandit Curves}
{app:bandit}

\appcontentsline{S}
{Detection Guarantees}
{app:guarantees}

\appcontentsline{T}
{High-Frequency Price Monitoring and the Limits of Aggregator Visibility}
{app:pricewatch}

\appcontentsline{U}
{Equivalence-Class Width: A TOST Check}
{app:tost}

\appcontentsline{V}
{Cost-Regret Scaling}
{app:regret}



\clearpage
\section{Limitations}
\label{app:limitations}
Our results are a measurement-driven view of the current open-weight provider
market, not a claim about a fully liquid token spot market of the kind studied for cloud compute
\citep{benyehuda2013spot}, where providers themselves arbitrage capacity
across preemptible instances \citep{miao2024spotserve,liu2025skyserve}. Across our coarse
multi-week measurement waves, listed prices move rarely, but high-frequency
monitoring shows that this apparent stability depends on sampling cadence and
provider-pool depth (Appendix~\ref{app:pricewatch}). The observed savings arise
mainly from cross-sectional provider dispersion at equal quality, while route
validity can change through availability, provider churn, quality drift, and
market re-election.
The light probes use small sample sizes and resolve large gaps more reliably than marginal ones. Rapid pinned probing can induce rate limits, and all requests traverse a single aggregator from one client location. This limits causal attribution: quality degradation may stem from quantization \citep{li2024quantized}, kernels, serving bugs \citep{kwon2023vllm}, sampling configuration, or aggregator interaction. Our claims are therefore attribution-agnostic: whatever the mechanism, the
endpoint-level behavior is observable by clients and matters for routing.
Likewise, the observed price--latency relationship is a client-visible
cross-sectional association rather than evidence that higher prices causally
produce faster service; Appendix~\ref{app:latency} reports the corresponding
robustness checks.
\textsc{facet} is also not a uniformly dominant online learner. It is designed for abrupt catastrophic cheap mines and task-selective feasibility. In tight-band cells with only marginal floor crossings, sliding-window learners may be cheaper or safer; under gradual ramps or multiple concurrent mines, the detector core is less favorable. We therefore present \textsc{facet} as an insurance mechanism with a construction-based safety property, not as a universal replacement for non-stationary bandits. Appendix~\ref{app:detector} reports the full detector-core comparison, and Appendix~\ref{app:drift} stress-tests this boundary under different drift shapes.
The robustness experiments also separate convenient assumptions from
minimum information requirements. \textsc{facet} tolerates imperfect task
assignment and sparse feedback, while symmetric label noise mainly induces
conservative fallback. Systematic evaluator bias is different: it can corrupt
certification unless ground-truth probes or audits provide an independent
quality signal. The system therefore still requires a maintained fallback,
some trustworthy evidence about served quality, and a mechanism that either
assigns a query to a facet or rejects it as unknown. In particular, our OOD
experiment evaluates the routing rule given an unknown-task decision; it does
not solve open-set task recognition itself. In the limit of no task signal
and no trustworthy quality evidence, safe certification is not possible.
The component-ablation and biased-evaluator stress tests use a single
Llama-3.3-70B/GSM8K exact-replay cell and should be interpreted as mechanism
and boundary tests rather than evidence that the same numerical effects hold
uniformly across all provider--task cells.

\section{Related Work Positioning}
\label{app:positioning}

Table~\ref{tab:positioning} makes explicit the decision axis studied in this paper. Existing client-side LLM routers---FrugalGPT, RouteLLM, CARROT, and MixLLM
\citep{chen2023frugalgpt,ong2024routellm,somerstep2025carrot,wang2025mixllm}---operate
primarily on the \emph{model-selection} axis: given a query, they decide whether to call a
stronger or weaker model \citep{sakota2024flyswat}, or how to cascade among models
\citep{yue2024cascades}. The table also lists a provider-side pricing mechanism and a cross-provider measurement leaderboard,
neither of which closes a measurement loop into client-side provider selection. Our setting is different and composable with those routers. We assume that a model has already been selected, and study the remaining question of which provider should serve the same open-weight model.
\begin{table}[h]
\centering
\small
\caption{\textbf{Positioning against representative routing and measurement work.}
Prior client-side routers select among models; our layer selects among
providers serving the same selected open-weight model.}
\label{tab:positioning}

\renewcommand{\arraystretch}{1.10}
\setlength{\tabcolsep}{7pt}

\begin{tabular}{@{}lccc@{}}
\toprule
& Decision axis & Cost model & Cross-provider quality \\
\midrule

FrugalGPT / RouteLLM \\ CARROT / MixLLM
& which \emph{model}
& static list price
& --- \\

PriLLM (provider-side)
& sets its own price
& endogenous
& --- \\

Token Arena (measurement)
& --- (leaderboard)
& ---
& static snapshot \\

\addlinespace[2pt]
\textbf{Ours}
& which \emph{provider}
& live, exogenous
& \textbf{measured, live} \\

\bottomrule
\end{tabular}

\end{table}
This distinction matters for evaluation. A model router does not decide among providers of the same model; it leaves that decision unmade. Conversely, our provider router does not decide whether a query should use an 8B, 70B, or frontier model. The two layers can be stacked: a model router first selects the model, then our provider router chooses the cheapest currently certified provider for that model and task.

Concurrent work extends model routing along axes that remain orthogonal to ours. Satisfaction-constrained routers optimize cost subject to a user-satisfaction target under limited feedback \citep{slarouter2026}; multi-turn routers condition the model choice on conversation history \citep{mtrouter2026}; and budget-paced adaptive routers handle non-stationary serving conditions. All three still choose \emph{which model} answers a query, and all three take the price of that choice from a posted list. Our layer is downstream of each of them: once a model is fixed, the price actually paid and the quality actually delivered are determined by which provider serves it, and that mapping must be measured rather than looked up.

\section{Detector-Core Comparison}
\label{app:detector}

The main paper presents \textsc{facet} as a deployment safety layer: it combines per-task feasibility certification, fail-safe anchoring, and slip detection. This appendix isolates the detector core in isolation from the certification and fail-safe layers. The purpose of this section is not to claim that the detector core is uniformly superior to all non-stationary bandits. Rather, it clarifies the regime where the detector core helps: abrupt, large-gap quality slips of a provider that is otherwise attractive by price.

We replay each benchmark cell as a piecewise-stationary stream. The stream begins with the measured segment, then the cheapest-safe provider slips below the floor, and later recovers, with a concurrent outage of the next-cheapest provider. We compare against a static argmin policy, cost-aware Thompson sampling, sliding-window UCB \citep{garivier2011switching}, and M-UCB \citep{cao2019mucb}, a change-detection bandit that restarts on detected reward shifts. All policies are evaluated against a clairvoyant cheapest-feasible oracle (Table~\ref{tab:scope}).

\begin{table}[h]
\centering
\small
\caption{\textbf{\textsc{facet}'s detector core versus non-stationary bandit baselines.}
Results cover 12 benchmark cells against a clairvoyant cheapest-feasible
oracle. Each safety metric is shown at the median and at each policy's own
worst cell. Cost premium is relative to the oracle; quality-regret is the
summed below-floor shortfall
$\sum_t \max(0,\theta-q_{a_t})$ per $10^3$ serves.}
\label{tab:scope}

\renewcommand{\arraystretch}{1.10}
\setlength{\tabcolsep}{7pt}

\begin{tabular}{@{}lrrrrr@{}}
\toprule
&
\multicolumn{1}{c}{Cost premium}
&
\multicolumn{2}{c}{Below-floor rate}
&
\multicolumn{2}{c}{Quality-regret} \\
\cmidrule(lr){2-2}
\cmidrule(lr){3-4}
\cmidrule(lr){5-6}

Policy
& vs.\ oracle
& Median
& Worst
& Median
& Worst \\
\midrule

Static-argmin
& $-6\%$
& $33.0\%$
& $33.0\%$
& $49.50$
& $49.50$ \\

Cost-Thompson
& $+4\%$
& $6.5\%$
& $47.6\%$
& $9.08$
& $20.45$ \\

SW-UCB
& $+13\%$
& $2.4\%$
& $56.6\%$
& $3.33$
& $6.26$ \\

M-UCB (change-detect)
& $+14\%$
& $2.5\%$
& $53.2\%$
& $3.71$
& $6.17$ \\

\addlinespace[2pt]
\rowcolor{gray!12}
\textbf{\textsc{facet} core}
& $+59\%$
& $\mathbf{1.3\%}$
& $42.6\%$
& $\mathbf{1.78}$
& $\mathbf{4.41}$ \\

\bottomrule
\end{tabular}

\end{table}
The detector core improves the median quality-regret in the abrupt-slip setting, but the improvement is conditional. Removing the own-baseline detector collapses safety because a slipping arm that dominates the serve stream is not caught by cohort comparison alone. Replacing floor-aware election with reward-maximizing election also increases below-floor serves, because a cheap sub-floor arm can still have high reward after the price term is included. Thus the detector and the floor-aware election address different failure modes.

This result should be read narrowly. The detector core is useful when the failure is an abrupt, large-gap slip of a provider that is attractive by price. It does not imply that \textsc{facet} dominates general non-stationary bandit methods. In tight-band cells, where arms differ only marginally around the floor, or under gradual drift, sliding-window learners can be cheaper or safer. For this reason, the main text uses the detector-core result only to support \textsc{facet}'s role as an insurance layer against catastrophic cheap mines, rather than as a general-purpose replacement for adaptive bandit routing.

\section{Measurement Protocol and Caveats}
\label{app:measurement}

All provider measurements are collected through a public multi-provider
aggregator. Each request is explicitly pinned to a single
provider and provider fallback is disabled, so the observed response reflects
the selected endpoint as delivered through the aggregator rather than an
aggregator-selected substitute.
For every model--task--provider cell, we record four client-visible quantities:
accuracy, realized per-call cost, mean latency, and availability. Availability
is measured as the fraction of requests successfully served under retry and
back-off. Failed requests are categorized as HTTP 429, 5xx, timeout, or
client-side failure. Realized cost is computed from the provider-reported or
aggregator-reported token charges associated with each request.

The light-probe suite covers multiple task forms, including multi-step
mathematical reasoning, information extraction, sentiment classification, and
code-output prediction. Code-output prediction is introduced in the third
measurement wave as an execution-verified generalization check. Because these
light probes use modest sample sizes, they are intended to resolve large
provider gaps rather than marginal differences of a few points. We therefore
supplement them with benchmark-scale GSM8K, MMLU, and HumanEval measurements
across all measured providers of the selected models.

Several caveats bound the interpretation of these measurements. First, rapid
provider-pinned probing can induce client-visible rate limits. We early-abort a
provider after repeated failures and report the served fraction as availability.
This may differ from global provider uptime, but it reflects the service
actually visible to a client using that route under the probing pattern.
Second, the measurement harness itself can introduce artifacts. In an early
run, an insufficient generation-token budget truncated reasoning outputs and
caused a strong model to appear severely degraded. We corrected the harness
before the reported experiments, but include this failure mode because
provider measurement is only meaningful when token budgets and decoding
settings are sufficient for the evaluated model.
Third, all measurements traverse one public aggregator and originate from one
client location. Latency differences can therefore reflect aggregator behavior
or geography in addition to provider-intrinsic performance. Accuracy is less
exposed to this confound because it concerns response content rather than
network path.
For these reasons, our measurement claims are deliberately attribution-agnostic.
We do not attempt to determine whether an observed quality difference is caused
by quantization \citep{frantar2023gptq,lin2024awq,li2024quantized}, kernels, batching
\citep{yu2022orca,kwon2023vllm}, sampling configuration, serving bugs, rate
limiting, or aggregator interaction. The routing problem only requires that
the endpoint-level behavior be observable to the client: regardless of its
internal cause, a provider that is currently slower, unavailable, or below the
required quality floor changes the routing decision.
\section{When Is the Safety Premium Worth Paying?}
\label{app:lambda}

The online policies in this paper trade two different quantities: monetary
serving cost and the number of below-floor answers. A policy that is safer but
more expensive cannot be declared preferable without specifying how much a
deployment values avoiding a below-floor serve.

Let $\lambda$ denote the application-level cost assigned to one below-floor
serve, expressed in dollars. We evaluate a policy using

\[
L
=
\mathrm{cost}
+
\lambda\cdot N_{\mathrm{below}},
\]

where $N_{\mathrm{below}}$ is the number of below-floor serves. For a baseline
policy $b$, \textsc{facet} becomes preferable when

\[
\lambda
>
\lambda^\star_b
=
\frac{
\mathrm{cost}_{\textsc{facet}}
-
\mathrm{cost}_{b}
}{
N_{\mathrm{below},b}
-
N_{\mathrm{below},\textsc{facet}}
}.
\]

We compute this break-even value over the benchmark replay cells with
$3000$ served queries per cell. In this comparison, \textsc{facet} incurs
\$1583 of total cost and 257 below-floor serves. The mean cost of an ordinary
served query is approximately \$0.53. Table~\ref{tab:lambda} reports the
resulting break-even thresholds.
\begin{table}[h]
\centering
\small
\caption{\textbf{Break-even cost of avoiding one below-floor serve.}
\textsc{facet} is preferred whenever the deployment-level loss associated
with one below-floor answer exceeds $\lambda^\star$. The final column
expresses the threshold in units of the mean price of an ordinary served
query.}
\label{tab:lambda}

\renewcommand{\arraystretch}{1.10}
\setlength{\tabcolsep}{7pt}

\begin{tabular}{@{}lrr@{\hspace{12pt}}rr@{}}
\toprule
&
\multicolumn{2}{c}{Policy outcome}
&
\multicolumn{2}{c}{Break-even threshold} \\
\cmidrule(lr){2-3}
\cmidrule(l){4-5}

Baseline
& Cost (\$)
& Below-floor serves
& $\lambda^\star$ (\$)
& Query-price equiv. \\
\midrule

Frozen map
& $1015$
& $990$
& $0.78$
& $1.5\times$ \\

Periodic $P{=}400$
& $1448$
& $595$
& $0.40$
& $0.8\times$ \\

M-UCB
& $1241$
& $314$
& $6.02$
& $11\times$ \\

SW-UCB
& $1230$
& $296$
& $9.08$
& $17\times$ \\

\bottomrule
\end{tabular}

\end{table}
The result distinguishes two regimes. Relative to a frozen provider map or
periodic full-pool refresh, \textsc{facet}'s additional cost is recovered once
avoiding a below-floor answer is valued at roughly the price of one ordinary
query. The comparison with adaptive sliding-window policies is less
one-sided. Against SW-UCB, the break-even value is approximately \$9.08, or
about $17$ ordinary query prices.
This threshold makes the safety--cost trade-off application dependent.
For throughput-oriented batch workloads in which a poor answer can simply be
retried, a sliding-window learner may be economically preferable. The balance
changes when a poor answer can propagate into a more costly action, such as
executing generated code, modifying persistent state, triggering a downstream
workflow, or requiring human review. We therefore do not claim that
\textsc{facet} universally dominates adaptive bandit policies. Its value
depends on the application-level consequence assigned to a below-floor serve.

\section{Price--Latency Trade-offs and Latency-Constrained Routing}
\label{app:latency}

The primary measured-map experiments minimize provider cost subject to
task-conditioned quality and health requirements, corresponding to
$\ell_{\max}=\infty$ in Section~\ref{sec:form}. Because the same measurements
also reveal substantial latency dispersion, we examine whether price contains
a systematic latency signal and how explicit latency requirements alter the
routing decision.
The analysis covers 23 model--task cells from 10 open-weight models, with
$100$--$150$ calls per endpoint. Across all latency analyses, 199 endpoint--run
regressions are available from 28,151 successful timed calls. The latest-run
correlation analysis contains 21 cells with complete measurements.

\begin{table}[h]
\centering
\small
\caption{\textbf{Within-cell Spearman correlations between provider price and service metrics.}
Results use the latest benchmark-scale measurement run. Negative
price--latency correlation indicates that higher-priced providers are faster.}
\label{tab:latency-cell-corr}

\renewcommand{\arraystretch}{1.08}
\setlength{\tabcolsep}{6pt}

\begin{adjustbox}{max width=\linewidth}
\begin{tabular}{@{}lrccc@{}}
\toprule
&
&
\multicolumn{3}{c}{\textbf{Spearman correlation with price}} \\
\cmidrule(lr){3-5}

Model
& $|P|$
& Latency
& Accuracy
& Availability \\
\midrule

\rowcolor{gray!12}
\multicolumn{5}{@{}l}{\emph{GSM8K}} \\

deepseek-chat-v3.1
& 8
& $-0.73$
& $+0.14$
& $+0.32$ \\

deepseek-v4-flash
& 16
& $-0.10$
& $+0.27$
& $+0.43$ \\

llama-3.1-8b-instruct
& 5
& $-0.50$
& $+0.22$
& $+0.35$ \\

llama-3.3-70b-instruct
& 11
& $-0.77$
& $-0.18$
& $-0.03$ \\

llama-4-maverick
& 4
& $-0.80$
& $-0.32$
& $+0.77$ \\

kimi-k2.6
& 16
& $-0.26$
& $+0.14$
& $0.00$ \\

gpt-oss-120b
& 18
& $-0.75$
& $-0.05$
& $+0.02$ \\

glm-5.2
& 27
& $-0.11$
& $+0.15$
& $-0.12$ \\

\addlinespace[3pt]
\rowcolor{gray!12}
\multicolumn{5}{@{}l}{\emph{HumanEval}} \\

deepseek-chat-v3.1
& 8
& $-0.80$
& $+0.15$
& $-0.18$ \\

gemma-3-27b-it
& 5
& $-0.30$
& $-0.40$
& $-0.05$ \\

llama-3.1-8b-instruct
& 5
& $-0.50$
& $+0.21$
& $0.00$ \\

llama-3.3-70b-instruct
& 11
& $-0.61$
& $-0.58$
& $-0.07$ \\

llama-4-maverick
& 5
& $-0.30$
& $-0.36$
& $+0.71$ \\

mistral-small-3.2-24b-instruct
& 3
& $-1.00$
& $+1.00$
& $-1.00$ \\

gpt-oss-120b
& 18
& $-0.81$
& $-0.36$
& $-0.12$ \\

\addlinespace[3pt]
\rowcolor{gray!12}
\multicolumn{5}{@{}l}{\emph{MMLU}} \\

deepseek-chat-v3.1
& 8
& $-0.82$
& $+0.05$
& $-0.04$ \\

gemma-3-27b-it
& 4
& $-0.80$
& $-0.11$
& $0.00$ \\

llama-3.1-8b-instruct
& 5
& $-0.50$
& $+0.40$
& $0.00$ \\

llama-3.3-70b-instruct
& 10
& $-0.39$
& $-0.75$
& $-0.05$ \\

llama-4-maverick
& 5
& $-0.80$
& $+0.20$
& $+0.35$ \\

mistral-small-3.2-24b-instruct
& 3
& $-0.50$
& $-0.50$
& $+0.50$ \\

\midrule

\textbf{Median}
& 21 cells
& $\mathbf{-0.61}$
& $+0.049$
& $0.00$ \\

\textbf{Negative cells}
& {}
& $\mathbf{21/21}$
& $10/21$
& $9/21$ \\

\bottomrule
\end{tabular}
\end{adjustbox}

\end{table}
Table~\ref{tab:latency-cell-corr} shows that the service dimensions behave
differently. Price has a strong and directionally consistent association
with latency: every one of the 21 cells has a negative correlation, with
median $\rho=-0.61$. Aggregating signs at the model level yields negative
price--latency association for all $10/10$ models ($p=0.002$ by a two-sided
sign test). Accuracy and availability do not show the same pattern. Their
median correlations are close to zero and their signs vary across cells.

\begin{table}[h]
\centering
\small
\caption{\textbf{Robustness of within-cell price correlations to measurement-run selection.}
The correlation direction and the number of cells with negative correlation
are reported for each service metric.}
\label{tab:latency-run-robust}

\renewcommand{\arraystretch}{1.10}
\setlength{\tabcolsep}{6pt}

\begin{tabular}{@{}lcc@{\hspace{8pt}}cc@{\hspace{8pt}}cc@{}}
\toprule
&
\multicolumn{2}{c}{\textbf{Latency}}
&
\multicolumn{2}{c}{\textbf{Accuracy}}
&
\multicolumn{2}{c}{\textbf{Availability}} \\
\cmidrule(lr){2-3}
\cmidrule(lr){4-5}
\cmidrule(lr){6-7}

Run selection
& Median $\rho$
& Negative cells
& Median $\rho$
& Negative cells
& Median $\rho$
& Negative cells \\
\midrule

Latest
& $\mathbf{-0.610}$
& $\mathbf{21/21}$
& $+0.049$
& $10/21$
& $+0.000$
& $9/21$ \\

Earliest
& $\mathbf{-0.610}$
& $\mathbf{21/23}$
& $-0.224$
& $14/23$
& $+0.000$
& $10/23$ \\

Across-run mean
& $\mathbf{-0.566}$
& $\mathbf{22/23}$
& $-0.063$
& $13/23$
& $+0.000$
& $11/23$ \\

\bottomrule
\end{tabular}

\end{table}
As shown in Table~\ref{tab:latency-run-robust}, the latency result is stable
to run selection: the median correlation remains between $-0.57$ and $-0.61$
and is negative in nearly every cell. The accuracy estimate, by contrast,
moves from $-0.224$ to $+0.049$ depending on the run-selection rule and is
never statistically significant. We therefore do not interpret the latest-run
$+0.049$ value as evidence of a positive price--accuracy relationship.
The result is not driven by a small set of specialized hardware providers.
After removing Groq, Cerebras, and SambaNova, the model-level median
price--latency correlation is $-0.706$, with all $10/10$ models still negative
($p=0.002$).
The single-aggregator measurement path remains a potential confound. Among
28,151 successful timed calls, the minimum observed end-to-end latency is
$0.08$ seconds, providing an upper bound on the shared client-to-aggregator
path for an individual request. Extreme rankings are also stable across
workloads: Groq ranks fastest in all six cells in which it appears, while
DigitalOcean ranks slowest in all of its observed cells.
As a stronger adversarial check, we add a fixed latency handicap $D$ to the
p95 latency of every non-cheapest provider. At $D=1$ and $2$ seconds, all
$17/17$ analyzed cells still contain a strictly faster quality-feasible
provider than the cheapest feasible route. The count remains $15/17$ at
$D=5$ seconds and $14/17$ at $D=10$ seconds.
These results support large speed differences rather than a precise ranking
among middle-tier providers. For example, normalized latency rank has standard
deviation $0.40$ for Nebius, $0.30$ for Parasail, and $0.26$ for DeepInfra
across observed cells. We therefore avoid claims about fine-grained provider
ordering.

To examine where the latency difference arises, we regress each call's latency
against its generated-token count. The slope approximates seconds per generated
token, while the intercept captures fixed request overhead.
\begin{table}[h]
\centering
\small
\caption{\textbf{Decomposition of the price--latency relationship.}
Correlations are computed across providers within each model--task cell.}
\label{tab:latency-mechanism}

\renewcommand{\arraystretch}{1.12}
\setlength{\tabcolsep}{9pt}

\begin{tabular}{@{}lcc@{}}
\toprule
Latency component
& Median $\rho$ with price
& Negative cells \\
\midrule

Seconds per generated token
& $\mathbf{-0.675}$
& $\mathbf{21/21}$ \\

Fixed request overhead
& $+0.274$
& $7/21$ \\

\bottomrule
\end{tabular}

\end{table}
Table~\ref{tab:latency-mechanism} shows that the association is concentrated
in generation throughput rather than fixed request overhead, which is
consistent with per-token generation speed being set by the serving
stack---batching and scheduling policy, kernels, and accelerator---rather than
by the request path \citep{yu2022orca,kwon2023vllm,agrawal2024sarathi}.
Within a cell, provider output length differs by a median of $1.35\times$ and
by at most $3.04\times$, whereas generation throughput differs by as much as
$18\times$. The latency gap is therefore not explained simply by cheaper
providers producing longer answers. Across 199 endpoint--run regressions, the
median $R^2$ is $0.64$, although $24\%$ of fits have $R^2<0.3$. We
consequently treat this decomposition as supporting mechanism evidence rather
than as a precise latency model or causal estimate.

The price--latency relationship has a direct consequence for the
cost-minimizing routing policy.
\begin{table}[h]
\centering
\small
\caption{\textbf{Latency consequence of choosing the cheapest quality-feasible provider.}}
\label{tab:latency-routing-penalty}

\renewcommand{\arraystretch}{1.10}
\setlength{\tabcolsep}{9pt}

\begin{tabular}{@{}lr@{}}
\toprule
Statistic & Value \\
\midrule

\rowcolor{gray!12}
\multicolumn{2}{@{}l}{\emph{Evaluation set}} \\

Cells with at least two feasible providers
& $17$ \\

\addlinespace[3pt]
\rowcolor{gray!12}
\multicolumn{2}{@{}l}{\emph{Latency consequence}} \\

Median normalized speed rank of cheapest feasible provider
& $1.00$ \\

Cheapest feasible provider is strictly slowest
& $\mathbf{9/17}$ \\

Median p95-latency penalty vs.\ fastest feasible provider
& $\mathbf{7.8\times}$ \\

p95-latency penalty range
& $2.3$--$63.1\times$ \\

\addlinespace[3pt]
\rowcolor{gray!12}
\multicolumn{2}{@{}l}{\emph{Price consequence}} \\

Median price premium of fastest feasible provider
& $\mathbf{1.92\times}$ \\

Price-premium range
& $1.41$--$5.56\times$ \\

\bottomrule
\end{tabular}

\end{table}
As summarized in Table~\ref{tab:latency-routing-penalty}, the matched-quality
savings in the main paper are not latency-neutral. Selecting the cheapest
feasible provider incurs a median $7.8\times$ p95-latency penalty relative to
the fastest feasible provider, while buying the fastest feasible endpoint
costs a median $1.92\times$ more.

We finally impose an explicit p95-latency deadline before minimizing cost.
\begin{table}[h]
\centering
\small
\caption{\textbf{Latency-constrained measured-map routing.}
Relative cost is normalized to the unconstrained cheapest quality-feasible
route. Route changes report the fraction of feasible cells whose selected
provider changes after imposing the deadline.}
\label{tab:latency-deadline}

\renewcommand{\arraystretch}{1.10}
\setlength{\tabcolsep}{9pt}

\begin{tabular}{@{}crrr@{}}
\toprule
&
\multicolumn{3}{c}{\textbf{Routing outcome}} \\
\cmidrule(lr){2-4}

Latency deadline
& Feasible cells
& Relative cost
& Route changes \\
\midrule

$30$ s
& $12/12$
& $1.00\times$
& $0\%$ \\

$10$ s
& $\mathbf{12/12}$
& $\mathbf{1.19\times}$
& $58\%$ \\

$5$ s
& $12/12$
& $1.31\times$
& $58\%$ \\

$2$ s
& $10/12$
& $2.03\times$
& $70\%$ \\

$1$ s
& $\mathbf{6/12}$
& $\mathbf{2.62\times}$
& $\mathbf{100\%}$ \\

\bottomrule
\end{tabular}

\end{table}
Table~\ref{tab:latency-deadline} shows that moderate latency requirements
preserve much of the cost advantage: a 10-second deadline leaves all $12/12$
evaluated cells feasible and raises cost by only $19\%$. More aggressive SLAs
shrink the feasible provider set. At a one-second deadline, only $6/12$ cells
remain feasible and the surviving routes cost $2.62\times$ the unconstrained
route.

These results clarify the role of latency in market-aware routing. Price
contains useful information about the cost--speed frontier, but it cannot
replace direct measurement of task-conditioned quality. The measured map can
therefore impose an application-specific latency requirement before minimizing
cost. We interpret the observed price--latency pattern as a cross-sectional,
client-visible association rather than as a causal effect of price, and our
single-aggregator measurements do not support claims about precise
provider-intrinsic latency rankings.

\section{Run-to-Run Variability}
\label{app:variability}

The replay results in the main paper average over multiple random seeds.
Because the point estimates across nearby probe rates are not monotone, we
separately examine whether these differences exceed ordinary run-to-run
variation.
Table~\ref{tab:iqr} reports the median and interquartile range over 20 seeds
for the probe-rate sweep under drift scenario S2.
\begin{table}[h]
\centering
\small
\caption{\textbf{Run-to-run variability of the \textsc{facet} probe-rate sweep.}
Results are computed over 20 seeds. The overlapping interquartile ranges at
low probe rates show that the ordering of nearby point estimates should not
be interpreted as a tuning effect.}
\label{tab:iqr}

\renewcommand{\arraystretch}{1.10}
\setlength{\tabcolsep}{8pt}

\begin{tabular}{@{}crr@{\hspace{10pt}}r@{}}
\toprule
&
\multicolumn{2}{c}{\textbf{Safety outcome}}
&
\multicolumn{1}{c}{\textbf{Monitoring cost}} \\
\cmidrule(lr){2-3}
\cmidrule(l){4-4}

Probe rate
& Below-floor median
& IQR
& Probe spend median (\$) \\
\midrule

$0.10$
& $0.00\%$
& [$0.00$, $0.80$]\%
& $17.0$ \\

$0.25$
& $0.44\%$
& [$0.00$, $0.80$]\%
& $17.6$ \\

$0.50$
& $0.60\%$
& [$0.00$, $0.88$]\%
& $28.2$ \\

$1.00$
& $0.14\%$
& [$0.00$, $0.96$]\%
& $29.8$ \\

$2.00$
& $0.92\%$
& [$0.52$, $1.36$]\%
& $48.9$ \\

$4.00$
& $1.02\%$
& [$0.52$, $1.40$]\%
& $47.5$ \\

\bottomrule
\end{tabular}

\end{table}
The four lowest probe-rate settings have strongly overlapping interquartile
ranges, all extending to zero. Their ordering therefore carries little
evidence of a systematic probe-rate effect, and we do not claim that one of
these operating points is optimal.

The more stable comparison is between monitoring regimes. Across the tested
settings, \textsc{facet}'s continuous monitoring keeps below-floor exposure
roughly in the $0$--$1.5\%$ range, whereas scheduled full-pool re-measurement
under the corresponding drift stress reaches substantially larger exposure.
The regime-level gap is much larger than the seed-to-seed variation within
\textsc{facet}. We therefore interpret Table~\ref{tab:reprobe-main} as evidence
for continuous monitoring versus periodic refresh, rather than as evidence for
fine-grained probe-rate tuning.
\section{Out-of-Sample Sensitivity to the Quality Floor and Health Threshold}
\label{app:theta}

The measured-map policy uses two deployment parameters: the quality margin
relative to the best measured provider in a model--task cell and the minimum
availability required for routing. The main experiments use a five-point
quality margin and a $90\%$ health threshold.
Evaluating these parameters in sample is insufficient because the same
measurements both define provider feasibility and evaluate the selected route.
As shown in Table~\ref{tab:theta-oos}, the resulting in-sample below-floor
rate is therefore $0\%$ for every threshold setting by construction. We
instead perform an out-of-sample sensitivity analysis using nine benchmark
cells for which two measurement waves with at least $100$ observations per
cell are available. The earlier wave is used to construct the routing map and
select a provider, while the later wave is used to evaluate whether that
provider still clears the corresponding quality floor.

\begin{table}[h]
\centering
\small
\caption{\textbf{Out-of-sample sensitivity to the quality margin and health threshold.}
Results cover nine benchmark cells. The earlier wave determines the route
and the later wave evaluates it. Values are below-floor rates.}
\label{tab:theta-oos}

\renewcommand{\arraystretch}{1.10}
\setlength{\tabcolsep}{7pt}

\begin{tabular}{@{}cc@{\hspace{8pt}}cc@{\hspace{8pt}}cc@{}}
\toprule
&
&
\multicolumn{2}{c}{\textbf{Measured-map router}}
&
\multicolumn{2}{c}{\textbf{OOS baselines}} \\
\cmidrule(lr){3-4}
\cmidrule(l){5-6}

Health threshold
& Cells
& In-sample
& OOS
& Cheapest
& Premium \\
\midrule

\rowcolor{gray!12}
\multicolumn{6}{@{}l}{\emph{Quality margin: 2 points}} \\

$80\%$
& $9$
& $0.0\%$
& $22.2\%$
& $44.4\%$
& $55.6\%$ \\

$90\%$
& $9$
& $0.0\%$
& $22.2\%$
& $44.4\%$
& $55.6\%$ \\

\addlinespace[3pt]
\rowcolor{gray!12}
\multicolumn{6}{@{}l}{\emph{Quality margin: 5 points}} \\

$80\%$
& $9$
& $0.0\%$
& $11.1\%$
& $22.2\%$
& $33.3\%$ \\

$90\%$
& $9$
& $0.0\%$
& $\mathbf{0.0\%}$
& $22.2\%$
& $33.3\%$ \\

\addlinespace[3pt]
\rowcolor{gray!12}
\multicolumn{6}{@{}l}{\emph{Quality margin: 10 points}} \\

$80\%$
& $9$
& $0.0\%$
& $\mathbf{0.0\%}$
& $0.0\%$
& $11.1\%$ \\

$90\%$
& $9$
& $0.0\%$
& $\mathbf{0.0\%}$
& $0.0\%$
& $11.1\%$ \\

\bottomrule
\end{tabular}

\end{table}

The out-of-sample evaluation removes the construction-induced zero of the
in-sample analysis. With a two-point quality margin, the measured-map router
falls below the later-wave floor in $22.2\%$ of cells; with a five-point
margin, the rate is $0.0$--$11.1\%$ depending on the health threshold; and
with a ten-point margin, it falls to $0.0\%$. Thus an aged map is not
guaranteed to preserve the feasibility observed when it was constructed.
The comparison with alternative provider-selection rules nevertheless remains
favorable across the sweep. At the two-point margin, measured-map,
cheapest-provider, and premium-provider routing have OOS below-floor rates of
$22.2\%$, $44.4\%$, and $55.6\%$, respectively. At the five-point margin,
the measured-map rate is $0.0$--$11.1\%$, compared with $22.2\%$ for cheapest
and $33.3\%$ for premium routing. At the ten-point margin, both measured-map
and cheapest routing reach $0.0\%$, while premium routing remains at
$11.1\%$. The measured-map policy is therefore never worse than either
comparison policy over the tested settings, while the nonzero OOS failures
also reinforce the need to update or certify a provider map as it ages.
\section{Why Provider Selection Has an Inverse-Bandit Structure}
\label{app:inverse-bandit}

In a classical stochastic bandit \citep{auer2002finite}, the reward of each arm is initially
unknown and must be learned through interaction. Same-model provider selection has a
different information structure. Provider prices, latency measurements, and client-visible health signals are
available before routing, while task-conditioned quality remains latent. The
unknown constraint is whether the current provider quality satisfies
\[
q_p(w,c,t)\ge\theta_{w,c}.
\]

This reverses the usual exploration problem. Rather than exploring arms to
discover which one offers the highest reward, the router first uses the
observed price ordering to identify attractive candidates and then spends
measurement effort to determine whether those candidates are feasible.
Probes and served-query feedback therefore supply evidence about an unknown,
time-varying constraint rather than about the economic objective itself.

This view is complementary to safe and constrained decision making
\citep{wu2016conservative,amani2019safe,pacchiano2021constrained,altman1999cmdp,achiam2017cpo,sui2015safeopt},
to budget-constrained bandits \citep{badanidiyuru2013bwk}, and to non-stationary and
change-detection bandits
\citep{besbes2014nonstationary,garivier2011switching,liu2018changedetection,cao2019mucb}. Those
frameworks provide mechanisms for constrained exploration or adaptation under
non-stationarity. Our setting additionally exposes the economic objective directly through
provider prices, while task-conditioned quality---and therefore the feasibility
constraint---remains latent. \textsc{facet} exploits this asymmetry by allowing price, latency, and health
to nominate candidate providers while requiring certification before an
uncertified candidate may receive user traffic.

\section{Drift-Shape Stress Tests}
\label{app:drift}

The guarantee in the main paper is matched to a specific drift shape: an abrupt, large-gap quality slip. This section reports stress tests outside that regime. We vary the drift pattern on the flagship cell and compare \textsc{facet}'s detector core to the best baseline under each condition (Table~\ref{tab:scope-drift}).
\begin{table}[h]
\centering
\small
\caption{\textbf{\textsc{facet}'s detector core across drift shapes.}
Values report below-floor rate and quality-regret on the flagship cell.
The detector is robust to frequent and correlated drift, but the advantage
does not hold under gradual ramps or multiple concurrent mines, which are
outside its abrupt-change assumption.}
\label{tab:scope-drift}

\renewcommand{\arraystretch}{1.10}
\setlength{\tabcolsep}{8pt}

\begin{tabular}{@{}lcc@{\hspace{10pt}}cc@{}}
\toprule
&
\multicolumn{2}{c}{\textbf{\textsc{facet} core}}
&
\multicolumn{2}{c}{\textbf{Best baseline}} \\
\cmidrule(lr){2-3}
\cmidrule(l){4-5}

Drift regime
& Below-floor
& q-regret
& Below-floor
& q-regret \\
\midrule

\rowcolor{gray!12}
\multicolumn{5}{@{}l}{\emph{Abrupt-change regimes}} \\

Abrupt, spaced (assumed)
& $1.3\%$
& $1.9$
& $1.4\%$
& $2.9$ \\

Frequent (sub-recovery)
& $1.3\%$
& $2.0$
& $2.5\%$
& $4.5$ \\

Correlated cohort slip
& $3.3\%$
& $3.9$
& $16.4\%$
& $21.0$ \\

\addlinespace[3pt]
\rowcolor{gray!12}
\multicolumn{5}{@{}l}{\emph{Outside the abrupt-change assumption}} \\

Gradual ramp
& $\mathbf{5.3\%}$
& $2.2$
& $1.3\%$
& $2.2$ \\

Multiple mines
& $3.0\%$
& $\mathbf{5.5}$
& $1.9\%$
& $4.0$ \\

\bottomrule
\end{tabular}

\end{table}
The stress test supports the limitation stated in the main paper. \textsc{facet} is strongest when the market contains a cheap provider that is either newly elected or abruptly becomes a large-gap mine. It is less suitable when degradation is gradual or when several arms become mines concurrently. In those regimes, sliding-window methods can detect smooth changes earlier or distribute exploration more favorably. This is why the main paper presents \textsc{facet} as a fail-safe certification layer for catastrophic provider mines, rather than as a replacement for all adaptive routing methods.
\section{Measured-Map Routing Results}
\label{app:offline-routing}

We compare the measured-map router against four practitioner baselines:
single-best, premium, random, and cheapest. Single-best selects the
highest-accuracy measured provider, premium selects the most expensive
provider, and cheapest ignores quality and routes purely by price.
\begin{table}[h]
\centering
\small
\caption{\textbf{Routing policy comparison over all measured model--task cells.}
Relative cost normalizes each cell to its cheapest provider. Below-floor is
the fraction of cells whose selected provider falls below the cell's quality
floor.}
\label{tab:baselines}

\renewcommand{\arraystretch}{1.10}
\setlength{\tabcolsep}{7pt}

\begin{tabular}{@{}lcccc@{}}
\toprule
Policy
& Mean acc.
& Rel.\ cost$^\ast$
& Below-floor
& Mean avail. \\
\midrule

Single-best (quality-first)
& $\mathbf{95\%}$
& $1.66\times$
& $\mathbf{0\%}$
& $96\%$ \\

Premium (most expensive)
& $92\%$
& $3.67\times$
& $17\%$
& $95\%$ \\

Random
& $92\%$
& $2.19\times$
& $22\%$
& $98\%$ \\

Cheapest (price-blind)
& $91\%$
& $\mathbf{1.00\times}$
& $28\%$
& $97\%$ \\

\addlinespace[2pt]
\rowcolor{gray!12}
\textbf{Ours}
& $94\%$
& $\mathbf{1.57\times}$
& $\mathbf{0\%}$
& $\mathbf{99\%}$ \\

\bottomrule
\end{tabular}

\end{table}
As shown in Table~\ref{tab:baselines}, because the same measurements define
feasibility and evaluate the policy, the measured-map router has zero
below-floor rate in sample by construction. We therefore also evaluate map
aging by selecting providers from an earlier measurement wave and scoring
them using a later wave.
\begin{table}[h]
\centering
\small
\caption{\textbf{Out-of-sample routing.}
Providers are selected using an earlier measurement wave and evaluated on a
later wave with $n{\ge}100$ benchmark accuracy.}
\label{tab:oos}

\renewcommand{\arraystretch}{1.10}
\setlength{\tabcolsep}{6pt}

\begin{tabular}{@{}lccc@{\hspace{12pt}}ccc@{}}
\toprule
&
\multicolumn{3}{c}{\textbf{In-sample}}
&
\multicolumn{3}{c}{\textbf{Out-of-sample}} \\
\cmidrule(lr){2-4}
\cmidrule(l){5-7}

Policy
& Accuracy
& Below-floor
& q-regret
& Accuracy
& Below-floor
& q-regret \\
\midrule

Single-best
& $87\%$
& $0\%$
& $0.0$
& $87\%$
& $0\%$
& $0.0$ \\

\rowcolor{gray!12}
\textbf{Router (ours)}
& $85\%$
& $0\%$
& $0.0$
& $86\%$
& $11\%$
& $0.0$ \\

Cheapest
& $84\%$
& $22\%$
& $0.1$
& $85\%$
& $22\%$
& $0.1$ \\

Premium
& $79\%$
& $44\%$
& $4.3$
& $80\%$
& $33\%$
& $4.0$ \\

\bottomrule
\end{tabular}

\end{table}
Table~\ref{tab:oos} shows that an aged map is no longer perfectly safe.
Its failures remain substantially less severe than blindly routing by price,
but the result motivates the online certification mechanism introduced in
Section~\ref{sec:facet}.
\section{Savings Beyond Saturated Cells}
\label{app:nonsaturated}

A potential concern with the matched-quality saving result is that several
light-probe cells are saturated: every measured provider obtains exactly the
same accuracy. In such cells, preserving measured quality is trivial, and a
large cost difference could therefore inflate the aggregate saving without
requiring the router to distinguish among providers of different quality.

We address this by partitioning the 18 measured cells into saturated cells,
where all providers have identical measured accuracy, and non-saturated cells,
where measurable cross-provider quality differences exist. We then recompute
the median saving of the cheapest measured-equivalent healthy provider
relative to the premium provider within each subset.
\begin{table}[h]
\centering
\small
\caption{\textbf{Matched-quality savings after separating saturated and
non-saturated model--task cells.}}
\label{tab:nonsaturated-summary}

\renewcommand{\arraystretch}{1.12}
\setlength{\tabcolsep}{10pt}

\begin{tabular}{@{}lrr@{}}
\toprule
Cell subset
& Cells
& Median saving \\
\midrule

All cells
& $18$
& $56.5\%$ \\

Saturated cells only
& $9$
& $56.5\%$ \\

\addlinespace[2pt]
\rowcolor{gray!12}
\textbf{Non-saturated cells only}
& $9$
& $\mathbf{50.0\%}$ \\

\bottomrule
\end{tabular}

\end{table}
As shown in Table~\ref{tab:nonsaturated-summary}, removing all nine saturated
cells reduces the median saving only from $56.5\%$ to $50.0\%$. The
matched-quality cost advantage is therefore not driven primarily by cells in
which provider quality is indistinguishable. Substantial savings remain when
providers exhibit measurable quality differences and the router must
explicitly exclude endpoints that fail the quality requirement.

Table~\ref{tab:nonsaturated-detail} reports all nine non-saturated cells.
\begin{table}[h]
\centering
\small
\caption{\textbf{Complete per-cell results for the nine non-saturated cells.}
Saving is the reduction of the selected measured-equivalent healthy route
relative to the premium provider.}
\label{tab:nonsaturated-detail}

\renewcommand{\arraystretch}{1.10}
\setlength{\tabcolsep}{9pt}

\begin{tabular}{@{}llrr@{}}
\toprule
Model
& Task
& Saving
& Providers \\
\midrule

llama-3.1-8b-instruct
& extraction
& $\mathbf{84.1\%}$
& $5$ \\

{}
& math
& $\mathbf{84.1\%}$
& $5$ \\

\addlinespace[3pt]
llama-3.3-70b-instruct
& extraction
& $79.8\%$
& $10$ \\

{}
& math
& $79.2\%$
& $11$ \\

\addlinespace[3pt]
llama-4-maverick
& extraction
& $50.0\%$
& $4$ \\

{}
& math
& $50.0\%$
& $4$ \\

\addlinespace[3pt]
deepseek-chat-v3.1
& extraction
& $45.7\%$
& $6$ \\

\addlinespace[3pt]
\rowcolor{gray!12}
gemma-3-27b-it
& math
& $\mathbf{0.0\%}$
& $4$ \\

\rowcolor{gray!12}
mistral-small-3.2-24b-instruct
& extraction
& $\mathbf{0.0\%}$
& $4$ \\

\bottomrule
\end{tabular}

\end{table}
The non-saturated cells also show that the saving is not uniform: individual
values range from $84.1\%$ to $0.0\%$. In particular,
\texttt{gemma-3-27b-it/math} and
\texttt{mistral-small-3.2-24b-instruct/extraction} yield no saving. These
zero-saving cases are consistent with the routing objective rather than
failures of it: when the cheapest endpoint does not satisfy the quality
requirement, the router must pay more for a feasible provider instead of
forcing a cheaper route. The $50.0\%$ median over the remaining
quality-differentiated cells therefore reflects savings available after
enforcing the measured quality constraint, rather than savings created by
saturated tasks.
\section{Periodic Re-Probing Baseline}
\label{app:reprobe}

We implement a deliberately simple alternative to online
certification. Every $P$ queries, the periodic policy probes every
provider, replaces its previous feasibility estimate with the newest
measurement, and then serves the cheapest provider whose latest
measurement clears the floor. Probe cost is charged to the policy.
Between refresh rounds, the policy does not receive new certification
evidence.

Table~\ref{tab:reprobe-full} evaluates this policy under three
representative drift regimes. S1 places a catastrophic cheap provider
in the market from the beginning, so periodic measurement can observe
the problem directly. S2 lets a previously certified cheap provider
slip during the stream. S3 uses a harder sequence of measured-style
provider changes.
\begin{table}[h]
\centering
\small
\caption{\textbf{Periodic re-probing under three drift regimes.}
S1 contains a cheap mine from the start; S2 lets a certified provider
slip mid-run; S3 contains repeated measured-style changes.}
\label{tab:reprobe-full}

\renewcommand{\arraystretch}{1.10}
\setlength{\tabcolsep}{7pt}

\begin{tabular}{@{}lr@{\hspace{10pt}}rrr@{}}
\toprule
&
\multicolumn{1}{c}{\textbf{Cost outcome}}
&
\multicolumn{3}{c}{\textbf{Safety outcome}} \\
\cmidrule(lr){2-2}
\cmidrule(l){3-5}

Policy
& Cost (\$)
& Below-floor
& Catastrophic
& q-regret / $10^3$ \\
\midrule

\rowcolor{gray!12}
\multicolumn{5}{@{}l}{\emph{S1: Cheap mine from the start}} \\

Periodic $P=200$
& $1834.8$
& $2.78\%$
& $2.78\%$
& $8.33$ \\

Periodic $P=500$
& $1212.8$
& $0.00\%$
& $0.00\%$
& $0.00$ \\

Periodic $P=1000$
& $1004.2$
& $0.00\%$
& $0.00\%$
& $0.00$ \\

SW-UCB
& $1040.3$
& $1.79\%$
& $1.79\%$
& $5.52$ \\

M-UCB
& $1073.2$
& $2.44\%$
& $2.44\%$
& $7.50$ \\

\textbf{\textsc{facet}}
& $1574.6$
& $0.00\%$
& $0.00\%$
& $0.00$ \\

\addlinespace[3pt]
\rowcolor{gray!12}
\multicolumn{5}{@{}l}{\emph{S2: Certified provider slips mid-run}} \\

Periodic $P=200$
& $1787.3$
& $5.00\%$
& $5.00\%$
& $15.00$ \\

Periodic $P=500$
& $1160.7$
& $5.00\%$
& $5.00\%$
& $15.00$ \\

Periodic $P=1000$
& $949.4$
& $3.89\%$
& $3.89\%$
& $11.67$ \\

SW-UCB
& $1035.5$
& $1.99\%$
& $1.99\%$
& $6.11$ \\

M-UCB
& $1056.8$
& $2.05\%$
& $2.05\%$
& $6.34$ \\

\textbf{\textsc{facet}}
& $1483.4$
& $\mathbf{0.44\%}$
& $\mathbf{0.44\%}$
& $\mathbf{1.32}$ \\

\addlinespace[3pt]
\rowcolor{gray!12}
\multicolumn{5}{@{}l}{\emph{S3: Repeated measured-style changes}} \\

Periodic $P=200$
& $1711.9$
& $13.89\%$
& $13.89\%$
& $20.83$ \\

Periodic $P=500$
& $1099.8$
& $17.47\%$
& $17.47\%$
& $26.21$ \\

Periodic $P=1000$
& $871.1$
& $30.11\%$
& $30.11\%$
& $45.17$ \\

SW-UCB
& $1018.9$
& $1.99\%$
& $1.99\%$
& $4.11$ \\

M-UCB
& $1047.2$
& $2.33\%$
& $2.33\%$
& $4.88$ \\

\textbf{\textsc{facet}}
& $1442.2$
& $\mathbf{1.14\%}$
& $\mathbf{1.14\%}$
& $\mathbf{1.70}$ \\

\bottomrule
\end{tabular}

\end{table}
The baseline is intentionally competitive. When the mine is already
present at a refresh point, as in S1, periodic measurement can be both
safe and cheaper than \textsc{facet}. Its weakness appears when a
provider changes state between refreshes. In S2 and S3, serving can
continue until the next global probe round even though the selected
endpoint has already fallen below the floor. This is the failure mode
summarized in Table~\ref{tab:reprobe-main} in the main paper.

\paragraph{Component control with the same served-feedback monitor.}
Because the component control is rerun over 40 seeds, the corresponding
periodic and \textsc{facet} reference values differ numerically from the
original operating-point sweep in Table~\ref{tab:reprobe-main}, while the
qualitative comparison remains unchanged.
The comparison above changes both the admission mechanism and the availability
of continuous served-feedback monitoring. To isolate certification, we attach
the identical LLR-CUSUM and class-margin monitor used by \textsc{facet} to the
periodic policy. Once the monitor triggers, the provider is removed until the
next scheduled full-pool re-measurement. This leaves scheduled admission in
place of \textsc{facet}'s certify-before-serve rule while holding the
post-admission monitor fixed.

We evaluate this control over 40 random seeds in two complementary regimes.
S-drift begins with a safe cheap provider that later degrades, isolating the
monitoring role. S-mine0 instead places the cheapest provider below the floor
from the outset, isolating admission.
\begin{table}[h]
\centering
\small
\caption{\textbf{Component ablation with a matched served-feedback monitor.}
Below-floor values are means with 95\% confidence intervals over 40 seeds.
S-drift isolates post-admission monitoring; S-mine0 isolates admission of an
initially unsafe provider.}
\label{tab:monitor-ablation}

\renewcommand{\arraystretch}{1.10}
\setlength{\tabcolsep}{7pt}

\begin{adjustbox}{max width=\linewidth}
\begin{tabular}{@{}lc@{\hspace{10pt}}rrr@{}}
\toprule
&
\multicolumn{1}{c}{\textbf{Safety outcome}}
&
\multicolumn{3}{c}{\textbf{Cost and fallback behavior}} \\
\cmidrule(lr){2-2}
\cmidrule(l){3-5}

Policy
& Below-floor
& Total cost (\$)
& Probe cost (\$)
& Anchor share \\
\midrule

\rowcolor{gray!12}
\multicolumn{5}{@{}l}{\emph{S-drift: Post-admission degradation}} \\

Periodic $P=400$
& $17.30\pm2.24\%$
& $1084$
& $487$
& $0.0\%$ \\

Periodic $P=400$ + monitor
& $0.45\pm0.16\%$
& $1349$
& $487$
& $3.6\%$ \\

Periodic $P=800$ + monitor
& $0.32\pm0.15\%$
& $1273$
& $278$
& $10.0\%$ \\

\textbf{\textsc{facet}} $r=0.1$
& $\mathbf{0.23\pm0.13\%}$
& $1391$
& $\mathbf{21}$
& $28.6\%$ \\

\textbf{\textsc{facet}} $r=1.0$
& $0.57\pm0.17\%$
& $1231$
& $\mathbf{35}$
& $15.8\%$ \\

\addlinespace[3pt]
\rowcolor{gray!12}
\multicolumn{5}{@{}l}{\emph{S-mine0: Initially unsafe provider}} \\

Periodic $P=400$
& $6.10\pm2.72\%$
& $1143$
& $487$
& $0.0\%$ \\

Periodic $P=400$ + monitor
& $0.17\pm0.07\%$
& $1443$
& $487$
& $5.1\%$ \\

Periodic $P=800$ + monitor
& $0.04\pm0.04\%$
& $1337$
& $278$
& $9.0\%$ \\

\textbf{\textsc{facet}} $r=0.1$
& $\mathbf{0.00\pm0.00\%}$
& $1458$
& $\mathbf{22}$
& $29.3\%$ \\

\textbf{\textsc{facet}} $r=1.0$
& $\mathbf{0.02\pm0.02\%}$
& $1177$
& $\mathbf{32}$
& $11.8\%$ \\

\bottomrule
\end{tabular}
\end{adjustbox}

\end{table}
Table~\ref{tab:monitor-ablation} separates the two roles. Under S-drift,
adding the same monitor to periodic admission removes nearly all of the safety
gap, because post-admission degradation is precisely the failure mode handled
by the monitor. Under S-mine0, certification provides the complementary
admission protection: an uncertified mine is not served while evidence is
gathered. Certification also changes measurement allocation substantially.
Periodic admission re-probes the full pool, whereas \textsc{facet} probes only
cheaper uncertified candidates, reducing dedicated probe spend by
$8$--$23\times$ in these operating points even though greater anchor use can
leave total cost similar.

\section{Benchmark-Scale Feasibility and Multi-Wave Drift}
\label{app:feasibility-drift}

The main text summarizes two properties of same-model provider feasibility:
it is task-selective and it changes over time. This section reports the
benchmark-scale provider measurements and the provider-level route changes
across measurement waves.

We evaluate GSM8K, MMLU, and HumanEval across every measured provider of all
models, yielding 93 endpoint-cells with $n{=}150$ for GSM8K and MMLU and
$n{=}100$ for HumanEval. Table~\ref{tab:bench} reports the full results.
\begin{table}[h]
\centering
\small
\caption{\textbf{Benchmark-scale verification across GSM8K, MMLU, and HumanEval.}
Most providers form quality-equivalent groups, but some deployments exhibit
task-selective degradation.}
\label{tab:bench}

\renewcommand{\arraystretch}{1.10}
\setlength{\tabcolsep}{6pt}

\begin{adjustbox}{max width=\linewidth}
\begin{tabular}{@{}lrcccc@{}}
\toprule
Model
& $|P|$
& Acc.\ range
& Cheapest (acc.\ @ \$/1M)
& Worst (acc.\ @ \$/1M)
& Sep. \\
\midrule

\rowcolor{gray!12}
\multicolumn{6}{@{}l}{\emph{GSM8K}} \\

DeepSeek-v3.1
& 8
& 94--96\%
& DeepInfra (96\% @ \$0.60)
& Mara (94\% @ \$1.15)
& --- \\

Gemma-3-27B
& 4
& 93--95\%
& DeepInfra (94\% @ \$0.12)
& Parasail (93\% @ \$0.27)
& --- \\

Llama-3.1-8B
& 5
& 76--85\%
& DeepInfra (79\% @ \$0.03)
& Novita (76\% @ \$0.03)
& --- \\

Llama-3.3-70B
& 11
& \textbf{56--96\%}
& DeepInfra (96\% @ \$0.21)
& Cloudflare (56\% @ \$1.27)
& $\mathbf{\checkmark}$ \\

Llama-4-Maverick
& 5
& 95--97\%
& DigitalOcean (96\% @ \$0.45)
& Google (95\% @ \$0.75)
& --- \\

Mistral-Small-24B
& 2
& 96--97\%
& Venice (97\% @ \$0.17)
& Parasail (96\% @ \$0.19)
& --- \\

\addlinespace[3pt]
\rowcolor{gray!12}
\multicolumn{6}{@{}l}{\emph{MMLU}} \\

DeepSeek-v3.1
& 6
& 85--91\%
& DeepInfra (85\% @ \$0.60)
& CoreWeave (85\% @ \$1.10)
& --- \\

Gemma-3-27B
& 4
& 78--80\%
& DeepInfra (78\% @ \$0.12)
& Parasail (78\% @ \$0.27)
& --- \\

Llama-3.1-8B
& 5
& 64--68\%
& DeepInfra (64\% @ \$0.03)
& DeepInfra (64\% @ \$0.03)
& --- \\

Llama-3.3-70B
& 11
& 78--84\%
& DeepInfra (81\% @ \$0.21)
& Cloudflare (78\% @ \$1.27)
& --- \\

Llama-4-Maverick
& 5
& 77--84\%
& DigitalOcean (83\% @ \$0.45)
& Google (77\% @ \$0.75)
& --- \\

Mistral-Small-24B
& 2
& 80--81\%
& DeepInfra (81\% @ \$0.14)
& Venice (80\% @ \$0.17)
& --- \\

\addlinespace[3pt]
\rowcolor{gray!12}
\multicolumn{6}{@{}l}{\emph{HumanEval}} \\

DeepSeek-v3.1
& 5
& 91--92\%
& AtlasCloud (91\% @ \$0.62)
& SiliconFlow (91\% @ \$0.64)
& --- \\

Gemma-3-27B
& 2
& 84--85\%
& Nebius (85\% @ \$0.20)
& Parasail (84\% @ \$0.27)
& --- \\

Llama-3.1-8B
& 5
& 59--70\%
& DeepInfra (59\% @ \$0.03)
& DeepInfra (59\% @ \$0.03)
& --- \\

Llama-3.3-70B
& 9
& \textbf{63--84\%}
& DeepInfra (84\% @ \$0.21)
& Cloudflare (63\% @ \$1.27)
& $\mathbf{\checkmark}$ \\

Llama-4-Maverick
& 4
& 85--89\%
& DeepInfra (89\% @ \$0.50)
& Google (85\% @ \$0.75)
& --- \\

\bottomrule
\end{tabular}
\end{adjustbox}

\end{table}
Many provider sets are tightly clustered in quality, supporting the
cheapest-equivalent routing regime used in the main paper. The important
exceptions are task-selective: the same deployment can be near the provider
field on one benchmark while falling far below it on another. The corresponding
equivalence analysis is reported separately in Appendix~\ref{app:tost}.
We also repeat the provider measurement process 13 and 43 days after the
initial collection. Across comparable cells, the selected route changes in
$4/17$ cells over the first interval and $2/18$ over the second. Table
~\ref{tab:flips} lists the provider-level transitions.
\begin{table}[h]
\centering
\small
\caption{\textbf{Decision changes across three measurement waves.}
Despite mostly stable prices, the cheapest feasible provider changes because
of health, availability, provider churn, and quality drift.}
\label{tab:flips}

\renewcommand{\arraystretch}{1.10}
\setlength{\tabcolsep}{7pt}

\begin{tabular}{@{}llll@{}}
\toprule
Cell
& Route before
& Route after
& Driver \\
\midrule

\rowcolor{gray!12}
\multicolumn{4}{@{}l}{\emph{Interval 1 (days 0--13)}} \\

deepseek-v3.1 / extraction
& AtlasCloud
& DeepInfra
& availability recovery \\

deepseek-v3.1 / math
& AtlasCloud
& DeepInfra
& availability recovery \\

llama-3.3-70B / extraction
& DeepInfra
& Nebius
& \textbf{quality slip below floor} \\

mistral-small / classification
& Venice
& DeepInfra
& rate limiting \\

\addlinespace[3pt]
\rowcolor{gray!12}
\multicolumn{4}{@{}l}{\emph{Interval 2 (days 13--43)}} \\

deepseek-v3.1 / classification
& DeepInfra
& AtlasCloud
& availability change \\

gemma-27B / math
& Phala
& DeepInfra
& \textbf{provider exit} \\

\bottomrule
\end{tabular}

\end{table}
The route changes are not primarily driven by rapid list-price movement.
Availability, rate limiting, provider membership, and quality can all change
while prices remain unchanged. This is why the main paper separates the
offline measured-map policy from the online certification problem handled by
\textsc{facet}.
\section{Assumption-Relaxation Experiments}
\label{app:assumptions}

The main text summarizes how \textsc{facet} behaves when the clean
information assumptions used in the primary replay are relaxed. This
section reports the individual sweeps. We separate three sources of
imperfect information: task assignment, quality-label reliability, and
feedback availability.

\paragraph{Task-label corruption.}
We retain the true task for scoring but randomly replace the task label
seen by the router with another known task. The context-blind baseline
pools all task evidence into a single provider certificate.
\begin{table}[h]
\centering
\small
\caption{\textbf{Task-label corruption.}
Exposure rises gradually as the task label becomes less reliable.}
\label{tab:task-error-full}

\renewcommand{\arraystretch}{1.12}
\setlength{\tabcolsep}{10pt}

\begin{tabular}{@{}lcc@{}}
\toprule
&
\multicolumn{2}{c}{\textbf{Catastrophic exposure}} \\
\cmidrule(lr){2-3}

Task-label error
& \textsc{facet}
& Context-blind \\
\midrule

$0\%$
& $0.00\%$
& $6.72\%$ \\

$5\%$
& $1.06\%$
& $6.72\%$ \\

$10\%$
& $2.23\%$
& $6.72\%$ \\

$20\%$
& $3.54\%$
& $6.72\%$ \\

$33\%$
& $\mathbf{4.72\%}$
& $\mathbf{6.72\%}$ \\

\bottomrule
\end{tabular}

\end{table}
Table~\ref{tab:task-error-full} shows gradual degradation as task-label
error increases. The $33\%$ point should not be interpreted as chance-level
classification for three tasks; a uniformly random three-way classifier
would have an error rate of approximately $67\%$. The experiment instead
measures the degradation over a substantial but still informative range of
task-label noise.

\paragraph{Unknown and OOD traffic.}
We next replace part of the stream with queries belonging to an unseen
task. The nearest-facet baseline forces the new task into an existing
facet. The fail-safe variant instead treats a query flagged as unknown
as uncertified.
\begin{table}[h]
\centering
\small
\caption{\textbf{Unknown-task traffic.}
Fail-safe routing avoids borrowing certification from an unrelated
known task.}
\label{tab:ood-full}

\renewcommand{\arraystretch}{1.12}
\setlength{\tabcolsep}{10pt}

\begin{tabular}{@{}lcc@{}}
\toprule
&
\multicolumn{2}{c}{\textbf{Catastrophic exposure}} \\
\cmidrule(lr){2-3}

OOD share
& Unknown $\rightarrow$ fail-safe
& Nearest known facet \\
\midrule

$10\%$
& $\mathbf{0.00\%}$
& $3.41\%$ \\

$25\%$
& $\mathbf{0.00\%}$
& $1.03\%$ \\

$50\%$
& $\mathbf{0.00\%}$
& $0.24\%$ \\

\bottomrule
\end{tabular}

\end{table}
Table~\ref{tab:ood-full} isolates the routing consequence of an available
reject signal. It does not claim that \textsc{facet} itself solves open-set
task detection.

\paragraph{Noisy quality judgments.}
To model an imperfect evaluator, we independently flip a fraction of
observed binary quality labels before they are passed to the
certification mechanism. True outcomes remain hidden from the router
and are used only for evaluation.
\begin{table}[h]
\centering
\small
\caption{\textbf{Symmetric quality-label noise.}
Random label flips primarily raise fallback cost rather than below-floor
exposure. Results average over 10 seeds.}
\label{tab:quality-noise-full}

\renewcommand{\arraystretch}{1.12}
\setlength{\tabcolsep}{10pt}

\begin{tabular}{@{}lcc@{}}
\toprule
&
\multicolumn{1}{c}{\textbf{Safety outcome}}
&
\multicolumn{1}{c}{\textbf{Cost outcome}} \\
\cmidrule(lr){2-2}
\cmidrule(l){3-3}

Judge-label error
& Below-floor
& Total cost (\$) \\
\midrule

$0\%$
& $0.28\%$
& $1130$ \\

$20\%$
& $\mathbf{0.00\%}$
& $2868$ \\

$30\%$
& $\mathbf{0.00\%}$
& $2942$ \\

\bottomrule
\end{tabular}

\end{table}
Table~\ref{tab:quality-noise-full} shows that the counter-intuitive fall in
exposure under symmetric noise should not be interpreted as evaluator
robustness. With $20\%$ random label flips, $95$--$97\%$ of traffic falls
back to the anchor, compared with roughly $12$--$16\%$ under clean feedback.
The near-zero exposure therefore reflects conservative abstention at
substantially higher cost.

\paragraph{Systematically biased quality judgments.}
Symmetric label noise penalizes strong and weak providers alike and therefore
tends to trigger fallback. A harder failure mode is an evaluator that
systematically over-credits a weak provider. We model targeted bias by changing
an incorrect answer from the mine, or from the provider that will later become
a mine, into a positive judgment with probability $b$. We additionally compare
three evidence channels: certification probes scored by the same judge,
certification probes scored against ground-truth answers, and ground-truth
probes combined with a 5\% independent audit of the currently served provider.
All results below use 40 random seeds.
\begin{table}[h]
\centering
\small
\caption{\textbf{Systematic evaluator bias and mitigation.}
Below-floor exposure under bias toward a weak provider. Parentheses report the
maximum over the 40 seeds where available.}
\label{tab:biased-judge}

\renewcommand{\arraystretch}{1.10}
\setlength{\tabcolsep}{7pt}

\begin{adjustbox}{max width=\linewidth}
\begin{tabular}{@{}lccc@{}}
\toprule
&
\multicolumn{3}{c}{\textbf{Below-floor exposure}} \\
\cmidrule(lr){2-4}

Bias
& Judged probes
& Gold probes
& Gold + 5\% audit \\
\midrule

\rowcolor{gray!12}
\multicolumn{4}{@{}l}{\emph{S-mine0: Initially unsafe provider}} \\

Targeted $30\%$
& $0.05\%$
& ---
& --- \\

Targeted $50\%$
& $0.79\%$ $(8.5)$
& $0.04\%$ $(1.3)$
& $\mathbf{0.01\%}$ \\

Targeted $70\%$
& $4.33\%$ $(37.7)$
& $0.31\%$ $(12.0)$
& $\mathbf{0.03\%}$ \\

Targeted $90\%$
& $72.6\%$
& $4.97\%$ $(99.4)$
& $\mathbf{0.43\%}$ $(9.3)$ \\

Lenient $90\%$
& $60.9\%$
& ---
& --- \\

\addlinespace[3pt]
\rowcolor{gray!12}
\multicolumn{4}{@{}l}{\emph{S-drift: Post-admission degradation}} \\

Targeted $30\%$
& $1.10\%$
& ---
& --- \\

Targeted $50\%$
& $2.39\%$
& $2.26\%$
& $\mathbf{1.13\%}$ \\

Targeted $70\%$
& $7.42\%$
& $9.16\%$
& $\mathbf{2.32\%}$ \\

Targeted $90\%$
& $37.4\%$
& $33.3\%$
& $\mathbf{7.12\%}$ \\

\bottomrule
\end{tabular}
\end{adjustbox}

\end{table}
Table~\ref{tab:biased-judge} makes the quality signal an explicit trust
boundary. With judged probes, cold-start certification remains reliable
under mild targeted bias but degrades sharply as the bias strengthens.
Ground-truth probes largely close this admission failure: through $70\%$
targeted bias, mean cold-start exposure remains at or below $0.31\%$. They
do not by themselves protect a provider that becomes bad after admission,
because drift detection operates on served feedback. Adding a 5\%
ground-truth audit supplies an independent monitoring signal and reduces
exposure by roughly $2$--$5\times$ in the tested biased-feedback regimes.
At $90\%$ targeted bias, the corresponding periodic baseline reaches
$93.9\%$ below-floor exposure in S-mine0 and $61.2\%$ in S-drift.

\paragraph{Sparse feedback.}
We then reveal quality feedback for only a random fraction of served
queries.
\begin{table}[h]
\centering
\small
\caption{\textbf{Sparse quality feedback.}
Safety degrades as labeled traffic becomes rare, but remains
substantially better than the comparison policies at the lowest label
rate. Results average over 10 seeds.}
\label{tab:sparse-full}

\renewcommand{\arraystretch}{1.12}
\setlength{\tabcolsep}{10pt}

\begin{tabular}{@{}lcc@{}}
\toprule
&
\multicolumn{2}{c}{\textbf{Below-floor exposure}} \\
\cmidrule(lr){2-3}

Observed-feedback fraction
& \textsc{facet}
& Reference \\
\midrule

$100\%$
& $0.28\%$
& --- \\

$50\%$
& $1.26\%$
& --- \\

$20\%$
& $1.47\%$
& --- \\

$5\%$
& $\mathbf{4.57\%}$
& SW-UCB: $\mathbf{13.32\%}$ \\

\bottomrule
\end{tabular}

\end{table}
As shown in Table~\ref{tab:sparse-full}, at a $5\%$ label rate the periodic
policy reaches substantially higher exposure in the same stress setting.
The key trend is that the certification mechanism degrades gradually as
observations become sparse rather than failing immediately once every query
is no longer labeled.

\paragraph{Joint stress test.}
Finally, we combine $20\%$ quality-label error with only $20\%$
observed feedback. This removes the clean-label and dense-feedback
assumptions simultaneously.
\begin{table}[h]
\centering
\small
\caption{\textbf{Joint information stress test.}
The combination of noisy and sparse feedback sharply reduces cost
efficiency, but does not produce a comparable collapse in safety.}
\label{tab:joint-full}

\renewcommand{\arraystretch}{1.12}
\setlength{\tabcolsep}{8pt}

\begin{tabular}{@{}lcccc@{}}
\toprule
&
\multicolumn{4}{c}{\textbf{Below-floor exposure}} \\
\cmidrule(lr){2-5}

Setting
& \textsc{facet}
& Periodic
& SW-UCB
& Static \\
\midrule

Clean
& $0.28\%$
& $14.30\%$
& $2.16\%$
& --- \\

$20\%$ noise $+$ $20\%$ labels
& $\mathbf{0.14\%}$
& $17.00\%$
& $10.52\%$
& $65.00\%$ \\

$20\%$ noise $+$ $20\%$ labels $+$ probe rate $0.25$
& $\mathbf{0.00\%}$
& $17.00\%$
& $10.52\%$
& $65.00\%$ \\

\bottomrule
\end{tabular}

\end{table}
Table~\ref{tab:joint-full} combines symmetric quality-label noise with sparse
feedback, relaxing the clean-label and dense-feedback assumptions
simultaneously. It should not be interpreted as removing every assumption.
Certification still requires some eventual quality evidence, a maintained
fallback, and a mechanism for assigning or rejecting query context. What the
experiment shows is narrower: clean labels and dense feedback are not
necessary for the observed fail-safe behavior, although losing them can
remove much of the economic advantage.
\section{Live Provider Deployment Details}
\label{app:live}

We complement the replay experiments with a continuous live run of
\textsc{facet} against real provider endpoints. The deployment uses
Llama-3.3-70B and 12 provider routes exposed through the same pinned
serving interface used by the measurement harness. Fresh GSM8K, MMLU,
and HumanEval queries are interleaved during the run.
Over approximately $36$ hours, the system serves 538 user queries and
issues 130 certification probes. Four calls fail at the API level. The
maintained anchor is priced at \$1.04/M tokens, while the cheap provider
eventually certified for most traffic is priced at \$0.21/M tokens.
Aggregate served accuracy is $88.3\%$.

The run begins conservatively. Over the first 100 queries, average
served price is \$0.924/M tokens because a large fraction of traffic
still falls back to the anchor. As certification evidence accumulates,
traffic moves toward the cheaper provider. After query 300, average
served price reaches \$0.210/M tokens, indicating that the cheap route
has become the dominant certified endpoint.
Serving traffic costs \$0.0635 and certification probes add \$0.0115.
Relative to serving the entire stream through the anchor, the resulting
reduction is $63.7\%$ for serving cost alone and $57.1\%$ after probe
cost is included.

The live run also exposed two implementation issues that are muted in
exact replay.First, certification based on very small samples produced false quarantines
on HumanEval. We therefore replaced the replay-style point-estimate rule with
Wilson-interval tests and increased the minimum certification sample to
$n_{\min}=20$. A provider is quarantined only when its Wilson upper bound
falls below the class-margin threshold. Second, an unavailable
endpoint repeatedly attracted probes because it remained cheap by
price, consuming a large fraction of the probing budget without
producing usable quality evidence. We therefore add availability
gating and exponential backoff before an endpoint can consume further
certification probes.

These changes do not alter the routing principle. They make explicit
that live certification depends not only on estimating quality but also
on managing sampling uncertainty and endpoint availability. No genuine
quality collapse occurs during the final clean deployment window, so
this experiment should be interpreted as evidence for live
certification and migration rather than as a real-world validation of
CUSUM slip detection.
\section{Composition with an Upstream Model Router}
\label{app:modelrouter}

The main paper treats provider selection as a layer beneath model
routing. We test this composition explicitly with a RouteLLM-style
difficulty router. The upstream policy first chooses between an 8B and
a 70B model, after which the provider policy chooses among endpoints
serving only the selected model. This experiment is a composability
stress test rather than an evaluation of a specific released RouteLLM
checkpoint.
\begin{table}[h]
\centering
\small
\caption{\textbf{Provider routing underneath model routing.}
Cost is reported over 3000 replayed queries. When the cheapest provider
is already safe, our layer leaves the route unchanged. When it is
degraded, provider-aware selection trades a small cost increase over
the cheapest default for substantially lower below-floor exposure.}
\label{tab:modelrouter-stack}

\renewcommand{\arraystretch}{1.10}
\setlength{\tabcolsep}{8pt}

\begin{tabular}{@{}lr@{\hspace{10pt}}ccc@{}}
\toprule
&
\multicolumn{1}{c}{\textbf{Cost outcome}}
&
\multicolumn{3}{c}{\textbf{Quality and safety outcome}} \\
\cmidrule(lr){2-2}
\cmidrule(l){3-5}

Provider policy
& Cost (\$)
& Accuracy
& Below-floor
& Catastrophic \\
\midrule

\rowcolor{gray!12}
\multicolumn{5}{@{}l}{\emph{GSM8K}} \\

Dearest
& $1435.7$
& $85.4\%$
& $24.6\%$
& $24.6\%$ \\

Cheapest
& $222.6$
& $92.5\%$
& $0.0\%$
& $0.0\%$ \\

\textbf{Ours}
& $222.6$
& $\mathbf{92.5\%}$
& $\mathbf{0.0\%}$
& $\mathbf{0.0\%}$ \\

\addlinespace[3pt]
\rowcolor{gray!12}
\multicolumn{5}{@{}l}{\emph{MMLU}} \\

Dearest
& $2054.2$
& $81.2\%$
& $44.1\%$
& $0.0\%$ \\

Cheapest
& $328.3$
& $81.2\%$
& $55.9\%$
& $0.0\%$ \\

\textbf{Ours}
& $336.7$
& $\mathbf{84.6\%}$
& $\mathbf{0.0\%}$
& $\mathbf{0.0\%}$ \\

\addlinespace[3pt]
\rowcolor{gray!12}
\multicolumn{5}{@{}l}{\emph{HumanEval}} \\

Dearest
& $2003.6$
& $77.1\%$
& $100.0\%$
& $42.5\%$ \\

Cheapest
& $319.7$
& $78.2\%$
& $57.5\%$
& $0.0\%$ \\

\textbf{Ours}
& $380.0$
& $\mathbf{85.3\%}$
& $\mathbf{0.0\%}$
& $\mathbf{0.0\%}$ \\

\bottomrule
\end{tabular}

\end{table}
Table~\ref{tab:modelrouter-stack} separates model selection from provider
selection. On GSM8K, the cheapest endpoint is already feasible and the
provider layer does nothing. On MMLU and HumanEval, the upstream model choice
alone does not prevent routing through degraded same-model endpoints. The
provider layer therefore addresses a decision that remains unresolved after
the upstream router has selected model capacity.
\section{Exact-Replay Bandit Curves}
\label{app:bandit}

\begin{figure}[h]\centering
\includegraphics[width=0.95\textwidth]{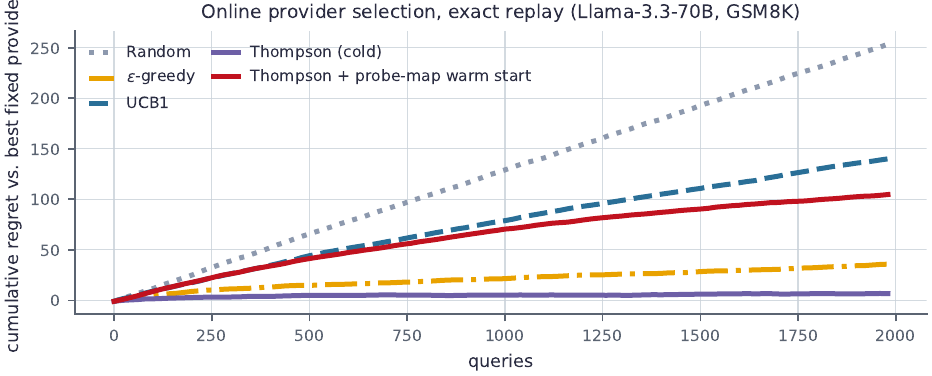}
\caption{Exact-replay online provider selection on Llama-3.3-70B/GSM8K. Cost-aware Thompson sampling approaches the best fixed provider within hundreds of queries. A probe-map warm start avoids early catastrophic exploration but can slow ranking if the prior is stale or task-form shifted.}
\label{fig:bandit}
\end{figure}

The main paper treats generic bandits as complementary to measurement, not as a replacement. Exact replay lets us test this cleanly because every provider's outcome on every benchmark item is recorded; replay therefore needs no off-policy correction. We simulate a query stream on the Llama-3.3-70B GSM8K cell, which contains 11 provider arms including the catastrophic 56\% endpoint (Figure~\ref{fig:bandit}).

The experiment shows two facts. First, online learning can fine-rank providers when enough query feedback is available. Second, generic bandits do not solve the safety problem exposed by our measurements. Their early behavior depends on whether the mine is structurally unattractive under the reward. In the measured cell, the catastrophic endpoint is also expensive, so cost-aware exploration avoids it partly by luck. If the catastrophic endpoint were cheap, an optimistic learner would explore it before detecting the failure. This motivates the fail-safe component of \textsc{facet}: uncertified arms are probed but not immediately served to user traffic.

\section{Detection Guarantees}
\label{app:guarantees}

This section gives the full certification and detector specification underlying
the safety discussion in the main paper. In replay, certification uses a
sliding window of the 200 most recent labelled observations with
$n_{\min}=12$. A facet is certified when its windowed accuracy is within
8 points of the best same-task provider with at least $n_{\min}$ observations
and at least $\theta-0.03$. The cohort test quarantines a facet when its
windowed accuracy falls more than 8 points below that cohort best. The anchor
is pre-certified and is not quarantined.
Certified facets are additionally monitored by a Bernoulli likelihood-ratio
CUSUM \citep{page1954,lorden1971,lai1998}. Its threshold is
$h=\log(1/0.01)$, the pre-change rate $\mu_s$ is the accuracy snapshot at
certification, and the post-change alternative is
$\min(\mu_s-0.10,\theta-0.02)$. Certification and monitoring therefore address
different failure modes: certification controls admission of an unverified
provider, whereas CUSUM detects a provider that was previously certified but
later degrades.

\begin{figure}[t]\centering
\includegraphics[width=\textwidth]{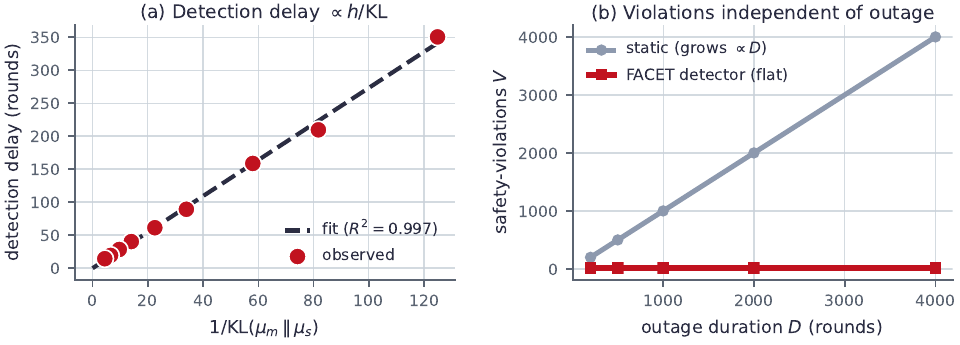}
\caption{Detector-core guarantees. \textbf{(a)} Detection delay scales linearly with
$1/\mathrm{KL}(\mu_m\Vert\mu_s)$ across the tested range of gap sizes ($R^2{=}0.997$).
\textbf{(b)} Below-floor serves remain flat as outage duration $D$ grows, because the provider is
quarantined after detection; a static policy accumulates violations throughout the outage.}
\label{fig:scope-out}
\end{figure}

\paragraph{Detection delay.}
For a slip from a safe rate $\mu_s$ to a mine rate $\mu_m=\mu_s-\Delta$, the expected CUSUM detection delay scales as
\[
\mathbb{E}[\tau]
\le
\frac{h}{\mathrm{KL}(\mu_m\Vert\mu_s)} + o(h),
\]
where $h$ is the threshold. For small gaps,
\[
\mathrm{KL}(\mu_m\Vert\mu_s)
=
\frac{\Delta^2}{2\mu_s(1-\mu_s)} + O(\Delta^3),
\]
so detection delay grows as $1/\Delta^2$ (Figure~\ref{fig:scope-out}(a)).

\paragraph{Outage-duration independence.}
Once a slipping provider is detected and quarantined, extending the duration of the outage does not add additional below-floor serves. Thus the safety cost depends on the detection delay, not on the length of the bad interval (Figure~\ref{fig:scope-out}(b)).
These plots support the detector component but should be interpreted within the scope of the paper. The fail-safe guarantee for newly elected uncertified arms does not rely on detecting a slip after user traffic has already been served; it follows from never serving uncertified facets. The CUSUM detector is needed for a different case: a facet that was correctly certified earlier but later degrades.

\section{High-Frequency Price Monitoring and the Limits of Aggregator Visibility}
\label{app:pricewatch}

The coarse multi-week measurements in the main text suggest relatively stable
listed prices. Here we test that observation at finer temporal resolution and
examine how much of the underlying price dynamics are visible through the
aggregator.

\paragraph{Setup.} Provider prices and availability are exposed by the aggregator's endpoint API,
which performs no inference and therefore costs nothing to poll. We recorded every provider's price
for four models every 15 minutes over a 24-hour window (388 observations), and logged a
\emph{re-election} whenever the cheapest provider for a model changed.

\paragraph{Result 1: apparent price stability depends on sampling cadence and pool depth.}
Table~\ref{tab:pricewatch} shows that the coarse-grained conclusion does not hold uniformly. Two
pools were indeed static. The two deeper pools were not: on a 30-provider pool we observed 14 price
changes and 3 re-elections in a single day, and on a 15-provider pool 2 price changes produced 2
re-elections. The re-elections were driven by ordinary competitive undercutting rather than by any
scheduled event---in one case a third-party host moved below the first-party endpoint, held the
lead for ten hours, and then gave it back.
Within this four-pool watch set, price movement is concentrated in the deeper provider pools: the pools with 8 and 12 providers did
not move at all, while those with 15 and 30 moved repeatedly. This both qualifies our own earlier
claim and strengthens the motivation for online re-certification: in a deep pool the identity of the
cheapest provider changes a few times per day, so a map refreshed on a daily or weekly cadence is
stale by construction. It also explains the discrepancy with the main text, whose cells are mostly
shallow pools observed at multi-week intervals.
\begin{table}[t]
\centering
\small
\caption{\textbf{High-frequency price monitoring over 24 hours at 15-minute resolution.}
A re-election is a change in the identity of the cheapest provider for that
model. Price movement and re-election frequency both increase with the number
of competing providers.}
\label{tab:pricewatch}

\renewcommand{\arraystretch}{1.10}
\setlength{\tabcolsep}{8pt}

\begin{tabular}{@{}lrccc@{}}
\toprule
&
&
\multicolumn{3}{c}{\textbf{Market dynamics}} \\
\cmidrule(lr){3-5}

Model
& Providers
& Price changes
& Re-elections
& Re-elections/day \\
\midrule

deepseek-chat-v3.1
& $8$
& $0$
& $0$
& $0.0$ \\

llama-3.3-70b-instruct
& $12$
& $0$
& $0$
& $0.0$ \\

\addlinespace[2pt]
deepseek-v4-pro-0813
& $15$
& $\mathbf{2}$
& $\mathbf{2}$
& $\mathbf{2.0}$ \\

deepseek-v4-flash-0731
& $30$
& $\mathbf{14}$
& $\mathbf{3}$
& $\mathbf{3.0}$ \\

\bottomrule
\end{tabular}

\end{table}

\paragraph{Result 2 (negative, pre-registered): the aggregator does not expose time-of-day pricing.}
One first-party provider in our watch list publishes an explicit peak/off-peak schedule: the peak
rate is twice the off-peak rate, and the peak window is a fixed weekday timetable. On one of the
models it serves, that provider's endpoint was the cheapest in the pool, and the gap to the next
host was $48\%$. Since the published peak multiplier is $100\%$, we registered the prediction ---
before the first peak window opened, and with the monitor already running --- that during peak hours
the first-party price should double and the cheapest provider should change to a third-party host,
reverting afterwards, on a published schedule covering $20.8\%$ of all hours.

The prediction was falsified. Across 28 observations inside peak windows the first-party price
reported by the aggregator was constant and identical to its off-peak value; no schedule-driven
re-election occurred. The aggregator evidently exposes a single contracted rate rather than passing
the provider's time-of-day tariff through to clients.

\paragraph{Result 3: the tariff is real at the source, and only the aggregator's view hides it.}
The falsified prediction leaves one alternative open: perhaps the published schedule is not applied
at all. We tested this directly against the provider's own API, bypassing the aggregator. The API
returns exact token usage but not price, so we inverted the billing: we issued matched batches of
requests---four batches whose output volume differs by less than $0.2\%$ ($17{,}969$, $17{,}987$,
$17{,}999$ and $18{,}000$ output tokens, with unique prompts so that no request hits the prompt
cache)---inside and outside peak windows, and read the account balance around each batch. Because
the batches are matched in volume, the ratio of the amounts charged is the ratio of the applied
rates.

The tariff is applied. The peak batch was charged substantially more than the off-peak batches for
the same work, and the peak charge agreed with the published peak table to within $1\%$
($\yen0.49$ observed against $\yen0.49$ tabulated). We do not report a precise multiplier, because
the provider settles charges against the balance with a lag of hours rather than minutes: a balance
that had been constant for ten minutes continued to decline three hours later, so charges cannot be
attributed to individual batches with the precision needed to separate, say, $1.7\times$ from
$2.0\times$. What the measurement does establish, and all we claim, is directional and unambiguous:
identical work costs materially more inside the peak window at the source, while the aggregator
reported an unchanging price throughout those same windows.

Taken together, Results 2 and 3 give the sharpest form of the point. The same provider, at the same
instant, presents two different prices depending on the channel through which it is reached: a
time-varying tariff to a direct client, and a single contracted rate to a client arriving through
an aggregator. Neither view is wrong; they are different market states. A router can only optimize
against the state it can observe, so the exogenous price in our formulation is channel-dependent.
For the aggregator-mediated client that this paper studies, the practical consequence is benign---
the tariff is invisible, so it can neither be exploited nor suffered---but it means that a client
that switched to direct access would face a materially different, and time-varying, cost landscape
than the one measured here.

We report this because the negative result is more informative than the positive one would have
been. It shows that even \emph{price}---the one component of provider state that is nominally
published---can be an incomplete view of the true market once it reaches a client through an
intermediary. Our entire argument is that a provider's serving quality must be measured rather than
read off a specification; this result extends that caution to the cost axis, and it bounds the
price-taker formulation of Section~\ref{sec:form}: the exogenous price a client optimizes against is
the price it can \emph{observe}, which need not be the price the provider actually charges at that
moment. Confirming the tariff at source would require direct first-party API access rather than
routing through an aggregator, and is a different experiment from the cross-provider comparison this
paper is built on: the phenomenon we study exists only where several providers serve the same
weights, which is precisely what an aggregator exposes and a single first-party API does not.

\section{Equivalence-Class Width: A TOST Check}
\label{app:tost}

A key empirical structure behind the measured-map router is that many provider sets form tight equivalence classes: several providers are close enough in quality that the cheapest healthy one can be selected without sacrificing task performance. However, this structure is not universal. We therefore test equivalence directly with a two-one-sided test (TOST) \citep{schuirmann1987} at a 5-point margin between the cheapest and best provider in each benchmark cell.
Equivalence is statistically established in \equivcells{} cells. This number
should be interpreted as a lower bound rather than as an estimate of how often
provider equivalence occurs. With approximately $n=150$ observations and
accuracy near $80\%$, the sampling uncertainty is comparable to or larger
than the five-point equivalence margin, making the test underpowered for
distinguishing small provider gaps. Failure to reject non-equivalence in the
remaining cells therefore does not establish that those providers are
meaningfully different.

The TOST result supports the narrower claim required by the routing analysis:
there exist same-model cells in which multiple providers are statistically
indistinguishable within the specified margin. In cells where equivalence
cannot be established, the router does not assume interchangeability; it
instead applies the explicit per-provider quality floor. The measured-map
policy is therefore floor-based rather than dependent on universal provider
equivalence.

\paragraph{Confidence-supported routing savings.}
The measured-map policy above applies its five-point margin to point
estimates. We therefore separately ask how much saving remains when sampling
uncertainty must also support the selected provider's quality gap. We consider
14 benchmark cells from the latest measurement wave with at least 100 items,
provider availability of at least $90\%$, and at least three available
providers per cell. For each candidate, we compute a one-sided 95\% Newcombe
(Wilson-score) upper confidence bound on its accuracy gap to the most accurate
provider and require that upper bound to be at most five percentage points.

Under the point-estimate rule, median saving relative to the premium provider
is $55\%$. Requiring the confidence interval to support the five-point margin
reduces this to $46\%$. For providers selected by the point-estimate rule, the
median upper confidence bound on the quality gap is $6.7$ points, showing that
$n\approx150$ items are often insufficient to certify a five-point margin.
Importantly, simply routing to the most accurate provider already saves
$45\%$. The statistically robust portion of the saving therefore comes
primarily from the fact that the most accurate provider is typically not the
most expensive one; exploiting the equivalence band provides additional
saving that requires larger samples to certify.

\section{Cost-Regret Scaling}
\label{app:regret}

The main safety guarantee concerns below-floor exposure. Cost-regret is a secondary quantity: when the router fails safe to an anchor or probes a candidate provider before admitting it, it can pay more than the clairvoyant cheapest-feasible oracle. Under spaced drift, each change triggers a bounded re-certification episode, so cost-regret should scale approximately linearly in the number of change-points.

We verify this on a controlled piecewise-stationary cell with eight arms. One cheap arm alternates between safe and mine states with fixed gap, and segment lengths are long enough for each change to resolve before the next. Median cost-regret over 12 seeds grows approximately linearly with the number of changes $L$. At $L\in\{1,2,4,8,16\}$, the per-change cost is nearly constant, with least-squares $R^2$ in the range $0.97$--$1.0$. Residual super-linearity at large $L$ comes from finite-sample recovery noise, such as a rehabilitated arm occasionally being re-quarantined.

This result is not the central guarantee of the paper. The load-bearing safety results are that newly elected uncertified arms are never served, and that slips of certified arms incur violations only until detection. The cost-regret scaling is reported for completeness because it characterizes the price paid for this safety.

\end{document}